\documentclass[10pt,twocolumn,letterpaper]{article}

\usepackage[pagenumbers]{iccv}

\usepackage{graphicx}
\usepackage{subcaption}
\usepackage{tikz}
\usepackage{multirow}
\usepackage{makecell}
\newcommand{\method}{{\color[RGB]{0,0,0}\texttt{VACE}}\xspace}
\newcommand{\methodbench}{{\color[RGB]{0,0,0}\texttt{VACE-Benchmark}}\xspace}
\newcommand{\methodbf}{{\color[RGB]{0,0,0}\textbf{\texttt{VACE}}}\xspace}
\definecolor{tabhighlight}{HTML}{e5e5e5}
\definecolor{video_quality}{HTML}{F0D695}
\definecolor{user_study}{HTML}{82ACD1}
\definecolor{score_average}{HTML}{98B567}

\definecolor{iccvblue}{rgb}{0.21,0.49,0.74}
\usepackage[pagebackref,breaklinks,colorlinks,allcolors=iccvblue]{hyperref}

\def\paperID{***}
\def\confName{ICCV}
\def\confYear{2025}

\title{VACE: All-in-One Video Creation and Editing}

\author{
Zeyinzi Jiang\thanks{Equal Contribution. $^\dagger$Project lead.} \quad Zhen Han$^*$ \quad Chaojie Mao$^*$$^\dagger$ \quad Jingfeng Zhang \quad Yulin Pan \quad Yu Liu \\
\\[-0.5em]
Tongyi Lab, Alibaba Group \\
}

\begin{document}

\maketitle

\begin{abstract}
\label{sec:abstract}
Diffusion Transformer has demonstrated powerful capability and scalability in generating high-quality images and videos. 
Further pursuing the unification of generation and editing tasks has yielded significant progress in the domain of image content creation.
However, due to the intrinsic demands for consistency across both temporal and spatial dynamics, achieving a unified approach for video synthesis remains challenging.
We introduce \textbf{VACE}, which enables users to perform \textbf{V}ideo tasks within an \textbf{A}ll-in-one framework for \textbf{C}reation and \textbf{E}diting.
These tasks include reference-to-video generation, video-to-video editing, and masked video-to-video editing. 
Specifically, we effectively integrate the requirements of various tasks by organizing video task inputs, such as editing, reference, and masking, into a unified interface referred to as the Video Condition Unit (VCU).
Furthermore, by utilizing a Context Adapter structure, we inject different task concepts into the model using formalized representations of temporal and spatial dimensions, allowing it to handle arbitrary video synthesis tasks flexibly.
Extensive experiments demonstrate that the unified model of VACE achieves performance on par with task-specific models across various subtasks. 
Simultaneously, it enables diverse applications through versatile task combinations.
%
Project page: \url{https://ali-vilab.github.io/VACE-Page/}.  
\end{abstract}

\section{Introduction}
\label{sec:intro}

In recent years, the domain of visual generation tasks has witnessed remarkable advancements, driven in particular by the rapid evolution of diffusion models~\cite{ddpm, sde, ddim, cfg, ldm, dit, unet}. 
Beyond the early foundational pre-trained models for text-to-image~\cite{sd3, pixart, hunyuan_dit} or text-to-video~\cite{ltx, wan2.1, goku} generation in the field, there has been a proliferation of downstream tasks and applications, such as repainting~\cite{sdinp,propainter}, editing~\cite{sdedit, ip2p, magicbrush, videocomposer, dreamVideo2}, controllable generation~\cite{controlnet, scedit},  frame reference generation~\cite{cogvideox, i2vadapter}, and ID-referenced video synthesis~\cite{largen, anydoor, consisid, phantom}. This array of developments highlights the dynamic and complex nature of the visual generation field.
To enhance task flexibility and reduce the overhead associated with deploying multiple models, researchers have begun to focus on constructing unified model architectures~\cite{ominictr, unireal} (\eg, ACE~\cite{ace, acepp} and OmniGen~\cite{omnigen}), aiming to integrate different tasks into a single image model, facilitating the creation of various application workflows while maintaining simplicity in usage.  
In the field of video, due to the collaborative transformations in both temporal and spatial dimensions, leveraging a unified model can present endless possibilities for video creation. 
However, leveraging diverse input modalities and ensuring spatiotemporal consistency are still chanlleging for unified video generation and editing.

We propose \methodbf, an all-in-one model for video creation and editing that performs tasks including reference-to-video generation, video-to-video editing, masked video-to-video editing, and free composition of these tasks, as illustrated in ~\cref{fig:show}.
On one hand, the aggregation of various capabilities reduces the costs of service deployment and user interaction. On the other hand, by combining the capabilities of different tasks within a single model, it addresses challenges faced by existing video generation models such as controllable generation of long videos, multi-condition and reference based generation, and continuous video editing, thereby empowering users with greater creativity. 
To achieve this, we utilize the current mainstream Diffusion Transformers (DiTs) structure as the foundational video framework and pre-trained text-to-video generation models~\cite{ltx, wan2.1}, which provides better basic capabilities and scalability for handling long video sequences. 
Specifically, \method takes into account the needs of different tasks during its construction and designs a unified interface, dubbed the Video Condition Unit (VCU), which integrates multiple modalities such as images or videos for editing, references, and masks.
Furthermore, to differentiate the visual modality information in editing and reference tasks, we introduce the concept decoupling strategy, enabling the model to understand what aspects need to be retained and what should be modified. 
Meanwhile, by employing a pluggable Context Adapter structure, concepts from different tasks (\eg, the areas or ranges of editing or reference) are injected into the model through collaborative spatiotemporal representation, enabling it to possess the capability of adaptive processing for unified tasks.

Due to the lack of existing multi-task benchmarks in video synthesis, we construct a dataset of 480 evaluation samples containing 12 different tasks, while evaluating the performance of the \method unified model by comparing it with existing specialized models. 
Experimental results demonstrate that our framework exhibits sufficient competitiveness in both quantitative and qualitative analyses. 
To the best of our knowledge, we are the first all-in-one model based on the video DiT architecture that concurrently supports such a wide range of tasks. 
Notably, this innovative framework allows for the compositional expansion of base tasks, enabling the construction of scenarios such as long video re-rendering, which provides a versatile and efficient solution for video synthesis, opening new possibilities for user-side video content creation and editing.

\begin{figure*}[!ht]  
    \centering  
    \includegraphics[width=0.99\textwidth]{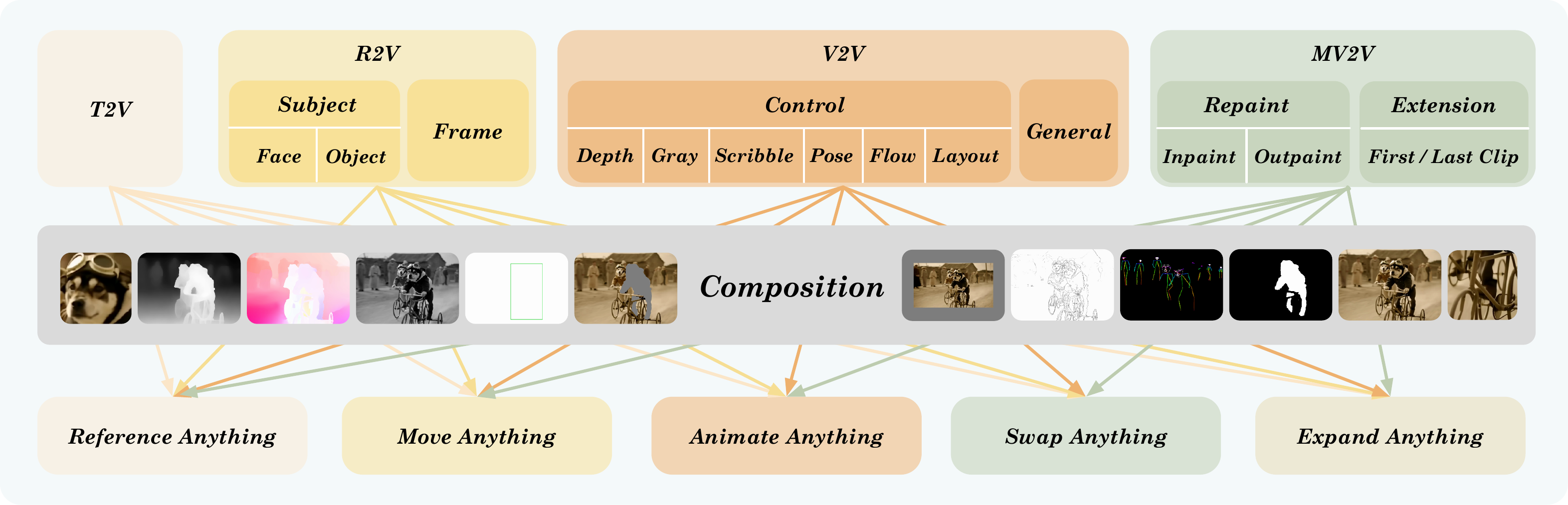}  
    \caption{\textbf{Task categories covered by \method.} Four basic tasks can be combined to create a vast number of possibilities.} 
    \label{fig:tasks} %
    \vspace{-8pt}
\end{figure*}

\section{Related Work}
\label{sec:related}

\noindent
\textbf{Visual Generation and Editing.}
With the rapid development of image~\cite{sd15, sd21, sdxl, sd3, pixart, flux} and video~\cite{i2vgen, cogvideox, ltx, hunyuanvideo} generation models, they are being used to create high-quality visual content and are widely applied in fields such as advertising, film special effects, game development, and animation production~\cite{midjourney, dalle3, wanx, gen3, hailuoai}. 
Meanwhile, to meet the diverse needs of visual media production and to enhance efficiency and quality, precise generation and editing methods have emerged. Models are required to perform generative creation based on multimodal inputs, such as depth, structure, pose, scene, and characters. 
According to the purposes of the input conditions, we can categorize them into two types: editing of the input and concept-guided re-creation. 
A significant portion of the work, such as ControlNet~\cite{controlnet}, ControlVideo, Composer~\cite{composer}, VideoComposer~\cite{videocomposer}, and SCEdit~\cite{scedit}, focuses on single-condition editing and multi-condition composite editing based on temporal and spatial alignment conditions. 
Additionally, some works that focus on interactive local editing scenarios, such as DragGAN~\cite{draggan} and MagicBrush~\cite{magicbrush}. 
Methods that guide generation based on semantic information from the input, such as Cone~\cite{cones}, Cone2~\cite{cones2}, InstantID~\cite{instantid}, and PuLID~\cite{pulid}, can achieve conceptual understanding of the input and inject it into the model for creative purposes.

\noindent
\textbf{Task-unified Visual Generative Model.}
As the complexity and diversity of user creations increase, relying solely on a single model or a complicated chain of multiple models can no longer provide a convenient and efficient path for implementing creative ideas. 
In image generation, a unified generation and editing framework has begun to emerge, allowing for more flexible creative approaches. Methods such as UltraEdit~\cite{ultraedit} and SEED-Data-Edit~\cite{seededit} provide general-purpose editing datasets, while techniques like InstructPix2Pix~\cite{ip2p}, MagicBrush~\cite{imagebrush}, and CosXL~\cite{cosxl} offer general instruction-based editing features. Additionally, methods like UniControl~\cite{unicontrol} and UNIC-Adapter~\cite{unicadapter} have unified controllable generation. Further advancements have led to the development of ACE~\cite{ace, acepp}, OmniGen~\cite{omnigen}, OmniControl~\cite{ominictr}, and UniReal~\cite{unireal}, which expand the scope of tasks by providing flexible controllable generation, local editing, and reference-guided generation.
In the video domain, due to the increased difficulty of generation, approaches often manifest as single-task single-model frameworks, offering capabilities for editing or reference generation, as seen in Video-P2P~\cite{videop2p}, MagicEdit~\cite{magicedit}, MotionCtrl~\cite{motionctrl}, Magic Mirror~\cite{magicmirror}, and Phantom~\cite{phantom}. \method aims to fill the gap for a unified model within the video domain, providing possibilities for complex creative scenarios.

\section{Method}
\label{sec:method}

\method is designed as a multimodal-to-video generation model, where text, image, video, and mask are integrated into a unified conditioning input.
To cover as many video generation and editing tasks as possible, we conduct in-depth research into existing tasks, then divide them into 4 categories according to their individual requirements of multimodal inputs.
Without losing generality, we specifically design a novel multimodal input format for each category under a Video Condition Unit (VCU) paradigm.
Finally, we restructure the DiT model for VCU inputs, making it a versatile model for a wide range of video tasks.

\subsection{Multimodal Inputs and Video Tasks.}

Despite existing video tasks being varying in complex user inputs and ambitious creative goals, we found that most of their inputs can be fully represented in 4 modalities: text, image, video, and mask.
Overall, as illustrated in ~\cref{fig:tasks}, we group these video tasks into 5 categories based on their requirements of these four multimodal inputs.

\begin{itemize}
\item \noindent\textbf{Text-to-Video Generation (T2V)}  is a basic video creation task and text is the only input.
\item \noindent\textbf{Reference-to-Video Generation (R2V)} requires additional images as reference inputs, making sure that specified contents, such as subjects of faces, animals and other objects, or video frames, appear in the generated video. 
\item \noindent\textbf{Video-to-Video Editing (V2V)} makes an entire change to a provided video, such as colorization, stylization, controllable generation, \etc. We use video control types whose control signals can be represented and stored as RGB videos, including depth, grayscale, pose, scribble, optical flow, and layout; however, the method itself is not limited to these.
\item \noindent\textbf{Masked Video-to-Video Editing (MV2V)} makes changes to an input video only within the provided 3D regions of interest (3D ROI), seamlessly blending in with the other unchanged regions, such as inpainting, outpainting, video extension, \etc. We use an extra spatiotemporal mask to represent the 3D ROI.
\item \noindent\textbf{Task Composition} includes all the compositional possibilities of the 4 types of video tasks above.
\end{itemize}

\subsection{Video Condition Unit}

\begin{table}[t]
\caption{
The formal representation of frames ($F$s) and masks ($M$s) under the four basic tasks. Frames and masks are aligned spatially and temporally.
}
\setlength\tabcolsep{15pt}
\begin{tabular}{c|c}
\toprule
Tasks  &  Frames ($F$s) \& Masks ($M$s) \\ \midrule 
\multirow{2}{*}{T2V}  
                      & $F=\{0_{h \times w}\} \times n$ 
\\
                      & $M=    \{1_{h \times w}\} \times n$ 
\\ \midrule 
\multirow{2}{*}{R2V}  
                      & $F=    \{r_1, r_2,...,r_l\} + \{0_{h \times w}\} \times n$ 
\\
                      & $M=    \{0_{h \times w}\} \times l + \{1_{h \times w}\} \times n$ 
\\ \midrule 
\multirow{2}{*}{V2V}  
                      & $F=\{u_1,u_2,...,u_n\}$                                                                  
\\
                      & $M=    \{1_{h \times w}\} \times n$
\\ \midrule 
\multirow{2}{*}{MV2V} 
                      & $F=\{u_1,u_2,...,u_n\}$                                                                  
\\
                      & $M=\{m_1,m_2,...,m_n\}$                                                                 
\\ 
\bottomrule
\end{tabular}
\vspace{-10pt}
\label{tab:vcu}
\end{table}

\begin{figure*}[t]  
    \centering  
    \includegraphics[width=0.99\textwidth]{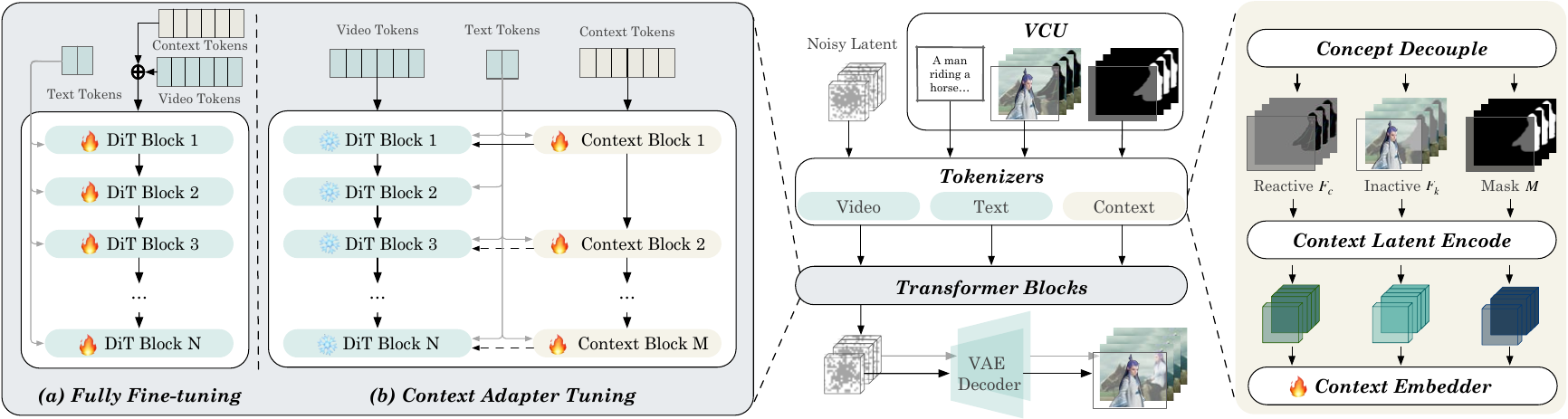}  
    \caption{\textbf{Overview of \method Framework.} Frames and masks are tokenized through Concept Decoupling, Context Latent Encode and Context Embedder. To achieve training with VCU as input, we employ two strategies, (a) Fully Fine-tuning and (b) Context Adapter Tuning. The latter converges faster and supports pluggable features.
    } 
    \label{fig:method} %
    \vspace{-8pt}
\end{figure*}

We propose an input paradigm, Video Condition Unit (VCU) to unify diverse input conditions into textual input, frame sequence, and mask sequence.
A VCU can be denoted as
\begin{equation}
   V=[T;F;M],
\end{equation}
where $T$ is a text prompt, while $F$ and $M$ are sequences of context video frames $\{u_1,u_2,...,u_n\}$ and masks $\{m_1,m_2,...,m_n\}$ respectively. Here, $u$ is in RGB space, normalized to \([-1, 1]\) and $m$ is binary, in which ``1"s and ``0"s symbolize where to edit or not. $F$ and $M$ are aligned both in spatial size $h \times w$ and temporal size $n$.
In T2V, no context frame or mask is required. 
To keep generality, we assign default value $0_{h \times w}$ to each $u$ denoting empty input, and set every $m$ to $1_{h \times w}$ meaning that all these 0-valued pixels are about to be re-generated. 
For R2V, additional reference frames $r_i$ are inserted in front of the default frame sequence, while all-zero masks $0_{h \times w}$ are inserted in front of the mask sequence. These all-zero masks mean that the corresponding frames should be kept unchanged. 
In V2V, context frame sequence is the input video frames and context mask is a sequence of $1_{h \times w}$. 
For MV2V, both context video and mask are required. 
The formal mathematical representations are shown in ~\cref{tab:vcu}.

VCU can also support task composition. For example, the context frames of reference-inpainting task are $\{r_1, r_2,...,r_l, u_1,u_2,...,u_n\}$ 
and the context masks are $\{0_{h \times w}\} \times l + \{m_1,m_2,...,m_n\}$.
In this case, users can modify $l$ objects in the video and regenerate based on the provided reference images. For another example, users only has a scribble image and wants to generate a video begining with the contents described by this scribble image, which is a scribble-based video extension task. The context frames are $\{u\}+\{0_{h \times w}\} \times (n-1)$ 
and the context masks are $\{1_{h \times w}\} \times n$. 
In this way, we can achieve multi-condition and reference control generation for long videos.

\subsection{Arichitecture}
We restructure the DiT model for \method, as shown in ~\cref{fig:method}, aiming to support multimodal VCU inputs.
Since there is an existing pipeline for text tokenization, we only consider about the tokenization of context frames and masks.
After tokenized, the context tokens combined with noisy video tokens and fine-tune the DiT model. 
Differ from that, we also propose a Context Adapter Tuning strategy, which allows context tokens to pass Context Blocks and added back to the original DiT Blocks. 

\subsubsection{Context Tokenization}

\noindent \textbf{Concept Decoupling.}
Two different visual concepts of natural video and control signals like depth, pose are encoded in $F$ simultaneously.
We believe that explicitly separating these data of different modalities and distributions is essential for model convergence.
The concept decoupling is based on masks and yields two frame sequences identical in shape: $F_c = F \times M$ and $F_k = F \times (1-M)$, 
where $F_c$ is called reactive frames contain all the pixels to be changed, while all the pixels to be kept are stored in $F_k$, named inactive frames.
Specifically, the reference images and the unchanged part of V2V and MV2V go to $F_k$, while control signals and those pixels about to change, such as gray pixels are collected to $F_c$. 

\noindent \textbf{Context Latent Encoding.}
A typical DiT processes noisy video latents $X\in \mathbb{R}^{n' \times h' \times w' \times d}$, where $n'$, $h'$ and $w'$ are the temporal and spatial shapes of the latent space.
Similar to $X$, $F_c$, $F_k$ and $M$ need to be encoded into a high-dimensional feature space to ensure the property of significant spatiotemporal correlations.
Therefore, we reorganize them together with $X$ into a hierachical and spatiotemporal aligned visual features.
$F_c$, $F_k$ are processed by video VAE and mapped into the same latent space of $X$, maintaining their spatiotemporal consistency. 
To aviod any unexpected mishmash of images and videos, reference images are separately encoded by VAE encoder and concatenated back along the temporal dimension, while the corresponding parts need to be removed during decoding.
$M$ is directly reshaped and interpolated.
After that, $F_c$, $F_k$, and $M$ are all mapping into latent spaces and are spatiotemporal aligned with $X$ in the shape of $n' \times h' \times w'$.

\noindent \textbf{Context Embedder.}
We extend the embedder layer by concatenating $F_c$, $F_k$ and $M$ in the channel dimension and tokenizing them into context tokens, which we refer to as the Context Embedder.
The corresponding weights to tokenize $F_c$ and $F_k$ is directly copied from the original video embedder, and the weights to tokenize $M$ are initialized by zeros.

\subsubsection{Fully Fine-Tuning and Context Adapter Tuning}
To achieve training with VCU as input, a simple methodology is fully fine-tuning the whole DiT model, as shown in ~\cref{fig:method}a.
Context tokens are added together with noisy tokens $X$, and all the parameters in DiT and the newly introduced Context Embedder will be updated during training.
To aviod fully fine-tuning and achieve faster convergence, as well as to establish a pluggable feature with the foundation model, we also propose another methodology processing the context token in a Res-Tuning~\cite{restuning} manner, as shown in ~\cref{fig:method}b.
Particularly, we select and copy several Transformer Blocks from the original DiT, forming a distributed and cascade-type Context Blocks. 
The original DiT processes video tokens and text tokens, while the newly added Transformer Blocks processes context tokens and text tokens.
The output of each Context Block is inserted back to the DiT blocks as an addictive signal, to assist the main branch in performing generation and editing tasks.
In this manner, the parameters of DiT are frozen. Only the Context Embedder and Context Blocks are trainable.
\section{Datasets}
\label{sec:datasets}

\begin{table*}[t]
\caption{
\textbf{Quantitative evaluations on \methodbench}. We compare the automated score metrics of the unified \method based on LTX-Video and the proprietary model on the dimensions of video quality and video consistency, as well as results of human user studies.
}

\footnotesize
\setlength{\tabcolsep}{2.5pt}
\begin{tabular}{l|l|ccccccccc|cccc}

\toprule

Type & Method & \multicolumn{9}{c|}{\textbf{Video Quality \& Video Consistency}} 
& \multicolumn{4}{c}{\textbf{User Study}} \\

& & \rotatebox{90}{\begin{minipage}{1.8cm} {\tikz\fill[video_quality] (0,0) circle (.5ex);} \centering Aesthetic Quality\end{minipage}}
& \rotatebox{90}{\begin{minipage}{1.8cm} {\tikz\fill[video_quality] (0,0) circle (.5ex);} \centering Background Consistency\end{minipage}}
& \rotatebox{90}{\begin{minipage}{1.8cm} {\tikz\fill[video_quality] (0,0) circle (.5ex);} \centering Dynamic Degree\end{minipage}}
& \rotatebox{90}{\begin{minipage}{1.8cm} {\tikz\fill[video_quality] (0,0) circle (.5ex);} \centering Imaging Quality\end{minipage}}
& \rotatebox{90}{\begin{minipage}{1.8cm} {\tikz\fill[video_quality] (0,0) circle (.5ex);} \centering Motion Smoothness\end{minipage}}
& \rotatebox{90}{\begin{minipage}{1.8cm} {\tikz\fill[video_quality] (0,0) circle (.5ex);} \centering Overall Consistency\end{minipage}}
& \rotatebox{90}{\begin{minipage}{1.8cm} {\tikz\fill[video_quality] (0,0) circle (.5ex);} \centering Subject Consistency\end{minipage}}
& \rotatebox{90}{\begin{minipage}{1.8cm} {\tikz\fill[video_quality] (0,0) circle (.5ex);} \centering Temporal Flickering\end{minipage}}
& \rotatebox{90}{\begin{minipage}{1.8cm} {\tikz\fill[score_average] (0,0) circle (.5ex);} \centering Normalized Average\end{minipage}}
& \rotatebox{90}{\begin{minipage}{1.8cm} {\tikz\fill[user_study] (0,0) circle (.5ex);} \centering Prompt Following\end{minipage}}
& \rotatebox{90}{\begin{minipage}{1.8cm} {\tikz\fill[user_study] (0,0) circle (.5ex);} \centering Temporal Consistency\end{minipage}} 
& \rotatebox{90}{\begin{minipage}{1.8cm} {\tikz\fill[user_study] (0,0) circle (.5ex);} \centering Video Quality\end{minipage}}
& \rotatebox{90}{\begin{minipage}{1.8cm} {\tikz\fill[score_average] (0,0) circle (.5ex);} \centering Average\end{minipage}} \\

\midrule 

I2V & I2VGenXL~\cite{i2vgen} & 55.20\% & 92.87\% & \textbf{60.00\%} & 63.31\% & 97.43\% & 23.78\% & 89.58\% & 95.67\% & 71.54\% & 2.65 & 1.60 & 2.34 & 2.20 \\
 & CogVideoX-I2V~\cite{cogvideox} & \textbf{57.78\%} & 94.80\% & 40.00\% & \textbf{68.23\%} & 98.69\% & 24.38\% & \textbf{93.84\%} & 97.84\% & 73.66\% & \textbf{3.30} & 2.28 & \textbf{3.19} & 2.92 \\
 & LTX-Video~\cite{ltx} & 56.12\% & 94.57\% & 35.00\% & 62.72\% & \textbf{99.27\%} & 24.92\% & 92.83\% & \textbf{98.41\%} & 72.89\% & 2.95 & 2.28 & 2.28 & 2.50 \\
 & \cellcolor{tabhighlight} \methodbf (Ours) & \cellcolor{tabhighlight} 57.53\% & \cellcolor{tabhighlight} \textbf{95.32\%} & \cellcolor{tabhighlight} 45.00\% & \cellcolor{tabhighlight} 68.03\% & \cellcolor{tabhighlight} 99.08\% & \cellcolor{tabhighlight} \textbf{25.13\%} & \cellcolor{tabhighlight} 93.61\% & \cellcolor{tabhighlight} 97.80\% & \cellcolor{tabhighlight} \textbf{74.38\%} & \cellcolor{tabhighlight} 3.20 & \cellcolor{tabhighlight} \textbf{4.00} & \cellcolor{tabhighlight} 2.54 & \cellcolor{tabhighlight} \textbf{3.24} \\

\midrule 

Inpaint & ProPainter~\cite{propainter} & 44.70\% & 95.64\% & \textbf{50.00\%} & \textbf{61.57\%} & 99.01\% & 18.48\% & 92.99\% & \textbf{98.47\%} & 70.15\% & 2.35 & \textbf{4.00} & \textbf{2.99} & \textbf{3.11} \\
& \cellcolor{tabhighlight} \methodbf (Ours) & \cellcolor{tabhighlight} \textbf{51.30\%} & \cellcolor{tabhighlight} \textbf{96.30\%} & \cellcolor{tabhighlight} \textbf{50.00\%} & \cellcolor{tabhighlight} 60.39\% & \cellcolor{tabhighlight} \textbf{99.12\%} & \cellcolor{tabhighlight} \textbf{21.12\%} & \cellcolor{tabhighlight} \textbf{94.59\%} & \cellcolor{tabhighlight} 98.21\% & \cellcolor{tabhighlight} \textbf{72.05\%} & \cellcolor{tabhighlight} \textbf{2.40} & \cellcolor{tabhighlight} \textbf{4.00} & \cellcolor{tabhighlight} 2.60 & \cellcolor{tabhighlight} 3.00 \\

\midrule 

Outpaint & Follow-Your-Canvas~\cite{followyourcanvas} & 53.30\% & 95.99\% & 5.00\% & \textbf{69.53\%} & 98.08\% & \textbf{25.90\%} & \textbf{95.38\%} & 97.20\% & 71.54\% & 3.05 & 2.00 & 1.63 & 2.23 \\
 & M3DDM~\cite{m3ddm} & 53.34\% & 95.87\% & \textbf{30.00\%} & 65.07\% & \textbf{99.22\%} & 25.43\% & 93.65\% & \textbf{98.85\%} & 73.16\% & 3.70 & 3.88 & 2.28 & 3.29 \\
 & \cellcolor{tabhighlight} \methodbf (Ours) & \cellcolor{tabhighlight} \textbf{57.04\%} & \cellcolor{tabhighlight} \textbf{96.55\%} & \cellcolor{tabhighlight} \textbf{30.00\%} & \cellcolor{tabhighlight} 69.49\% & \cellcolor{tabhighlight} 99.20\% & \cellcolor{tabhighlight} 25.36\% & \cellcolor{tabhighlight} 94.47\% & \cellcolor{tabhighlight} 98.47\% & \cellcolor{tabhighlight} \textbf{74.25\%} & \cellcolor{tabhighlight} \textbf{3.90} & \cellcolor{tabhighlight} \textbf{3.92} & \cellcolor{tabhighlight} \textbf{3.58} &  \cellcolor{tabhighlight} \textbf{3.80} \\

\midrule 

Depth & Control-A-Video~\cite{controlavideo} & 50.62\% & 91.71\% & \textbf{70.00\%} & \textbf{67.76\%} & 97.58\% & 24.48\% & 88.10\% & 96.58\% & 72.35\% & 2.70 & 2.28 & 1.54 & 2.17 \\
 & VideoComposer~\cite{videocomposer} & 50.03\% & 94.18\% & \textbf{70.00\%} & 59.44\% & 96.23\% & 24.95\% & 89.79\% & 94.38\% & 70.74\% & 2.60 & 2.44 & 2.17 & 2.40 \\
 & ControlVideo~\cite{controlvideo} & \textbf{63.30\%} & 95.02\% & 10.00\% & 65.13\% & 96.49\% & 24.20\% & 92.29\% & 95.42\% & 70.07\% & 2.55 & 2.50 & 1.82 & 2.29 \\
 & \cellcolor{tabhighlight} \methodbf (Ours) & \cellcolor{tabhighlight} 56.72\% & \cellcolor{tabhighlight} \textbf{96.12\%} & \cellcolor{tabhighlight} 60.00\% & \cellcolor{tabhighlight} 66.41\% & \cellcolor{tabhighlight} \textbf{98.84\%} & \cellcolor{tabhighlight} \textbf{25.27\%} & \cellcolor{tabhighlight} \textbf{94.09\%} & \cellcolor{tabhighlight} \textbf{97.27\%} & \cellcolor{tabhighlight} \textbf{74.99}\% & \cellcolor{tabhighlight} \textbf{3.10} & \cellcolor{tabhighlight} \textbf{3.92} & \cellcolor{tabhighlight} \textbf{2.66} & \cellcolor{tabhighlight} \textbf{3.23} \\

\midrule 

Pose & Text2Video-Zero~\cite{text2videozero} & 57.63\% & 87.67\% & \textbf{100.00\%} & \textbf{70.74\%} & 79.65\% & 23.94\% & 84.82\% & 76.57\% & 59.69\% & 2.15 & 2.00 & 1.88 & 2.01 \\
 & ControlVideo~\cite{controlvideo} & \textbf{65.37\%} & 94.56\% & 25.00\% & 65.28\% & 97.32\% & 25.19\% & 92.76\% & \textbf{96.82\%} & 72.45\% & 2.15 & 1.80 & 2.03 & 1.99 \\
 & Follow-Your-Pose~\cite{followyourpose} & 48.79\% & 86.80\% & \textbf{100.00\%} & 67.41\% & 90.12\% & 26.10\% & 80.18\% & 88.02\% & 66.43\% & 2.00 & 2.60 & 1.58 & 2.06 \\
 & \cellcolor{tabhighlight} \methodbf (Ours) & \cellcolor{tabhighlight} 60.17\% & \cellcolor{tabhighlight} \textbf{94.92\%} & \cellcolor{tabhighlight} 75.00\% & \cellcolor{tabhighlight} 64.71\% & \cellcolor{tabhighlight} \textbf{98.63\%} & \cellcolor{tabhighlight} \textbf{26.44\%} & \cellcolor{tabhighlight} \textbf{94.82\%} & \cellcolor{tabhighlight} 96.60\% & \cellcolor{tabhighlight} \textbf{76.13\%} & \cellcolor{tabhighlight} \textbf{2.95} & \cellcolor{tabhighlight} \textbf{3.96} & \cellcolor{tabhighlight} \textbf{2.63} & \cellcolor{tabhighlight} \textbf{3.18} \\

\midrule 

Flow & FLATTEN~\cite{flatten} & \textbf{56.23\%} & 95.80\% & 70.00\% & 61.65\% & 97.86\% & \textbf{26.23\%} & 93.94\% & 96.17\% & 74.42\% & \textbf{3.50} & 2.40 & \textbf{3.19} & 3.03 \\
 & \cellcolor{tabhighlight} \methodbf (Ours) & \cellcolor{tabhighlight} 55.76\% & \cellcolor{tabhighlight} \textbf{96.07\%} & \cellcolor{tabhighlight} \textbf{75.00\%} & \cellcolor{tabhighlight} \textbf{65.37\%} & \cellcolor{tabhighlight} \textbf{98.98\%} & \cellcolor{tabhighlight} 25.89\% & \cellcolor{tabhighlight} \textbf{94.63\%} & \cellcolor{tabhighlight} \textbf{96.93\%} & \cellcolor{tabhighlight} \textbf{75.90\%} & \cellcolor{tabhighlight} 2.90 &\cellcolor{tabhighlight}  \textbf{3.75} & \cellcolor{tabhighlight} 2.60 & \cellcolor{tabhighlight} \textbf{3.08} \\

\midrule 

R2V & Keling1.6~\cite{klingai} & 62.13\% & 96.04\% & \textbf{85.00\%} & 69.27\% & 99.38\% & \textbf{27.82\%} & 93.79\% & 97.79\% & \textbf{78.81\%} & \textbf{4.22} & \textbf{4.10} & 3.80 & \textbf{4.04} \\
 & Pika2.2~\cite{pika} & 62.48\% & 96.79\% & 65.00\% & 69.87\% & 99.37\% & 26.02\% & 95.93\% & 98.90\% & 77.87\% & 4.00 & 3.85 & \textbf{3.87} & 3.91 \\
 & Vidu2.0~\cite{vidu} & \textbf{64.30\%} & 96.85\% & 35.00\% & 67.03\% & \textbf{99.66\%} & 26.53\% & 96.73\% & \textbf{99.41\%} & 76.47\% & 3.90 & 3.85 & 3.77 & 3.84 \\
 & \cellcolor{tabhighlight} \methodbf (Ours) & \cellcolor{tabhighlight} 63.25\% & \cellcolor{tabhighlight} \textbf{98.03\%} & \cellcolor{tabhighlight} 30.00\% & \cellcolor{tabhighlight} \textbf{72.29\%} & \cellcolor{tabhighlight} 99.51\% & \cellcolor{tabhighlight} 25.85\% & \cellcolor{tabhighlight} \textbf{98.54\%} & \cellcolor{tabhighlight} 99.15\% & \cellcolor{tabhighlight} 76.76\% & \cellcolor{tabhighlight} 3.47 & \cellcolor{tabhighlight} 3.42 & \cellcolor{tabhighlight} 3.30 & \cellcolor{tabhighlight} 3.40 \\

\bottomrule
\end{tabular}
\label{tab:comp}
\vspace{-8pt}
\end{table*}

\subsection{Data Construction}

To obtain an all-in-one model, the diversity and complexity of the required data construction also increase. Existing common text-to-video and image-to-video tasks only require constructing pairs of text and video. However, for the tasks in \method, the modalities need to be further expanded to include target videos, source videos, local masks, reference and more. To efficiently and rapidly acquire data for various tasks, it is imperative to maintain video quality while also conducting instance-level analysis and understanding of the video data. 

To achieve this, we first analyze the video data itself by performing shot slicing and preliminarily filtering out data based on resolution, aesthetic score, and motion amplitude. Next, we label the first frame of the video using RAM~\cite{ram} and combine it with Grounding DINO~\cite{groundingdino} for detection, utilizing the localization results to perform secondary filtering on videos with target areas that are either too small or too large. Furthermore, we employ the propagate operation of SAM2~\cite{sam2} for video segmentation to obtain instance-level information across the video. Leveraging the results of video segmentation, we filter instances in the temporal dimension by calculating the effective frame ratio based on the mask area threshold.

In the actual training process, the construction for different tasks also needs to be tailored according to the characteristics of each task: 
\textbf{1)} For some controllable video generation tasks, we pre-extract depth~\cite{midas}, scribble~\cite{infordraws}, pose~\cite{openpose, dwpose}, and optical flow~\cite{raft} from the filtered videos. For gray and layout tasks, we create data on the fly.
\textbf{2)} For repainting tasks, random instances from the videos can be masked for inpainting, while the inverse of the mask enables the construction of outpainting data. Augmentation of the masks~\cite{lama} allows for unconditional inpainting.
\textbf{3)} In the case of extension tasks, we extract key frames such as the first frame, last frame, frames from both ends, random frames, and segments from both ends to support a wider variety of extension types.
\textbf{4)} For reference tasks, we can extract several face or object instances from the videos and apply offline or online augmentation operations to create paired data.
Notably, we randomly combine all the previously mentioned tasks for training to accommodate a broader range of model application scenarios. Additionally, for all operations involving masks, we perform arbitrary augmentation to satisfy various granular local generation requirements.

\subsection{VACE-Benchmark}

Significant progress has been made in the field of video generation. However, a scientific and thorough evaluation of the performance of these models remains an urgent issue that needs to be addressed. VBench~\cite{vbench} and VBench++~\cite{vbenchpp} have established a precise evaluation framework for text-to-video and image-to-video tasks through an extensive assessment suite and dimensional design. Nevertheless, as the video generation ecosystem continues to evolve, more derivative tasks have begun to emerge, such as video reference generation and video editing, for which a comprehensive benchmark is still lacking. To address this gap, we propose \methodbench to evaluate various downstream tasks related to video in a systematic manner.

Starting from the data sources, we recognize that real videos and generated videos may exhibit different performance characteristics during evaluation. Thus, we collected a total of 240 high-quality videos categorized by their sources, encompassing various data types, including text-to-video, inpainting, outpainting, extension, grayscale, depth, scribble, pose, optical flow, layout, reference face, and reference object tasks, with an average of 20 samples for each task. The input modalities include input videos, masks, and reference, and we also provide the original videos to enable further processing by developers based on the specific characteristics of each task. Regarding the data prompts, we supply the original captions of the videos for quantitative assessment, as well as rewritten prompts tailored to the specific tasks to evaluate the models' creativity. 
\section{Experiments}
\label{sec:experiments}

\subsection{Experimental Setup}

\begin{figure*}[!ht]  
    \centering  
    \includegraphics[width=0.99\textwidth]{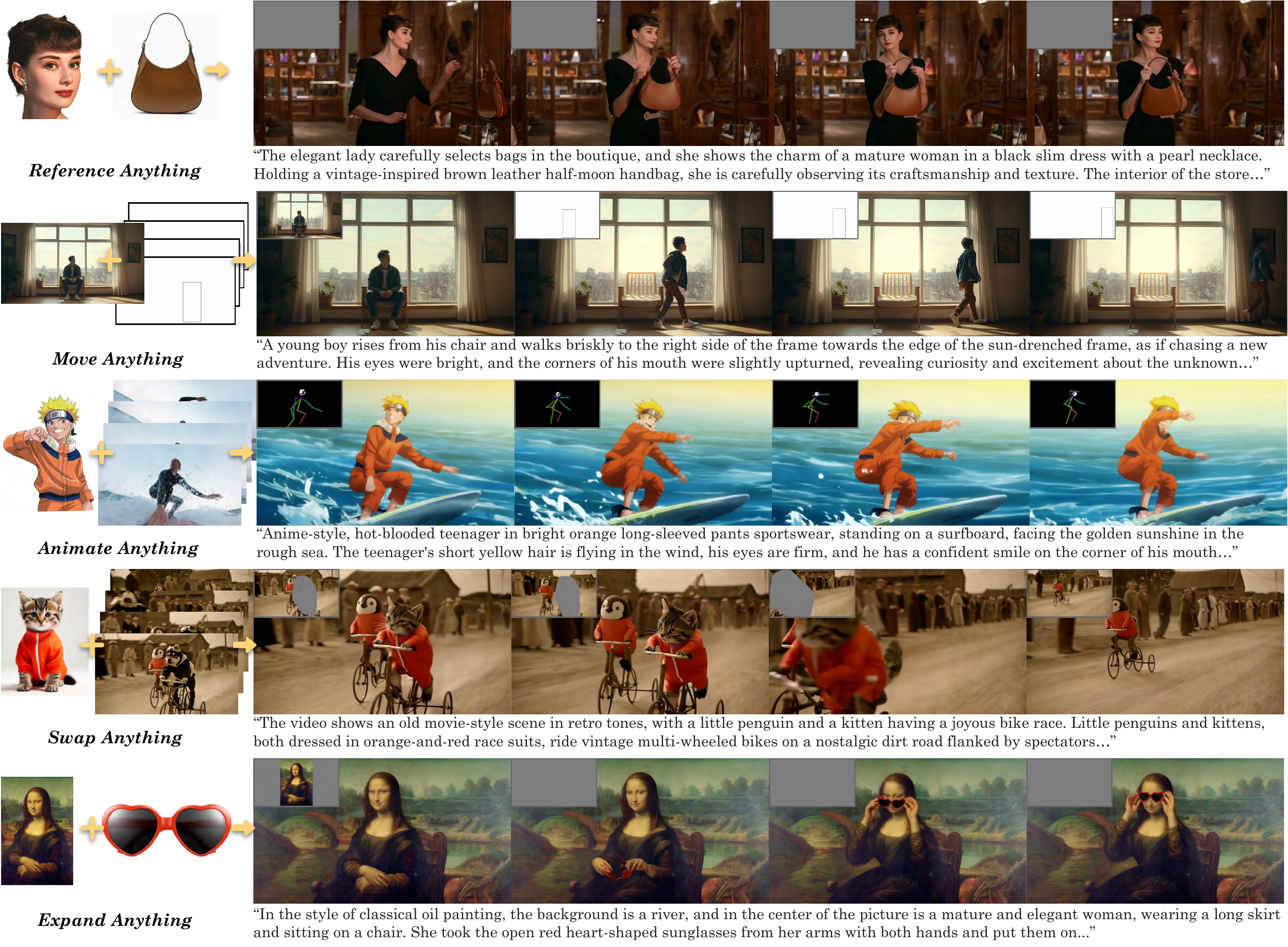}
    \caption{\textbf{Visualization results of compositional tasks.} \method creatively enables reference-, move-, animate-, swap-, and expand-anything.} 
    \label{fig:cases} %
\end{figure*}


\noindent
\textbf{Implementation Details.} 
\method is trained based on Diffusion Transformers for text-to-video generation at various scales. It utilizes LTX-Video-2B~\cite{ltx} for faster generation, while Wan-T2V-14B~\cite{wan2.1} is used specifically for higher-quality outputs, supporting resolutions of up to 720p. 
The training employs a phased approach. Initially, we focus on foundational tasks such as inpainting and extension, which are considered modal complementary to the pre-trained text-to-video models. This includes the incorporation of masks and the learning of contextual generation in both spatial and temporal dimensions. Next, from a task expansion perspective, we progressively transition from single input reference frames to multiple input reference frames and from single tasks to composite tasks. Finally, we fine-tune the model's quality using higher-quality data and longer sequences.
The input for model training accommodates arbitrary resolutions, dynamic durations, and variable frame rates to support diverse input needs of users.

\noindent
\textbf{Baselines.} 
Our goal is to achieve the unification of video creation and editing tasks, and currently, there is no comparable all-in-one video generation model available, which leads us to focus our evaluation on comparing our general model with proprietary task-specific models.
Moreover, due to the numerous tasks involved and the lack of open-sourced methods for many of them, we conduct our comparisons on models that are available either offline or online.
Specifically for the tasks, we compare the following: 
\textbf{1)} For the I2V task, we examine I2VGenXL~\cite{i2vgen}, CogVideoX-I2V~\cite{cogvideox}, and LTX-Video-I2V~\cite{ltx}; 
\textbf{2)} In the repainting task, we compare the ProPainter~\cite{propainter} for removal inpainting, while Follow-Your-Canvas~\cite{followyourcanvas} and M3DDM~\cite{m3ddm} are compared for outpainting; 
\textbf{3)} For controllable task, we use Control-A-Video~\cite{controlavideo}, VideoComposer~\cite{videocomposer}, and ControlVideo~\cite{controlvideo} under depth conditions, and compare Text2Video-Zero~\cite{text2videozero}, ControlVideo~\cite{controlvideo}, and Follow-Your-Pose~\cite{followyourpose} under pose conditions, as well as FLATTEN~\cite{flatten} under optical flow conditions; 
\textbf{4)} In reference generation, given the absence of open-sourced models, we compare commercial products Keling1.6~\cite{klingai}, Pika2.2~\cite{pika}, and Vidu2.0~\cite{vidu}.

\noindent
\textbf{Evaluation.} 
To comprehensively evaluate the performance of various tasks, we employ the \methodbench for assessment. Specifically, we divide the evaluation into automatic scoring and a user study for manual assessment. 
For the automatic scoring, we utilize select metrics from VBench~\cite{vbench} to assess video quality and video consistency, including eight indicators: aesthetic quality, background consistency, dynamic degree, imaging quality, motion smoothness, overall consistency, subject consistency, and temporal flickering. 
For the manual assessment, we utilize the mean opinion score (MOS) as our evaluation metric, focusing on three aspects: prompt following, temporal consistency, and video quality. In practice, we anonymize the generated data and randomly distribute it to different participants for scoring on a scale from 1 to 5.

\subsection{Main Results}

\noindent
\textbf{Quantitative Evaluation.}
We compare \method comprehensive model based on LTX-Video with task proprietary approaches on \methodbench. 
For certain tasks, we follow existing methods; for example, although we support generating based on any frame, we conduct comparisons using the first-frame reference approach from current open-source methods to ensure fairness.
From ~\cref{tab:comp}, we can seen that for the tasks of I2V, inpainting, outpainting, depth, pose, and optical flow, our method demonstrates better performance than other open-source methods across eight indicators of video quality and video consistency, with normalized average metrics showing superior results. 
Some competing methods can only generate at a resolution of 256, have very short generation durations, and exhibit instability in temporal coherence, resulting in poorer performance on automatic metric calculations. 
For the R2V task, there is still a certain gap in metrics compared to commercial models for a small-scale model that aims for fast generation, while being comparable to the metrics of Vidu 2.0.
According to the results of human user studies, our method consistently performs better in evaluation metrics across multiple tasks, aligning well with user preferences.

\noindent
\textbf{Qualitative Results.}
In ~\cref{fig:show}, we present the results of the \method single model across various tasks. It is evident that the model achieves a high level of performance in video quality and temporal consistency.
Furthermore, in composition tasks shown in ~\cref{fig:cases}, our model showcases impressive abilities, effectively integrating different modalities and tasks to produce results that cannot be generated by existing single or multiple models, thereby demonstrating its strong potential in the fields of video generation and editing. For example, in the ``Move Anything'' case, by providing a single input image and a movement trajectory, we are able to precisely move the characters in the scene with specified direction while maintaining coherence and narrative consistency.

\subsection{Ablation Studies}

\begin{figure}[t]
    \centering
    \vspace{4pt}
    \begin{subfigure}[b]{0.48\columnwidth}
        \includegraphics[width=\textwidth]{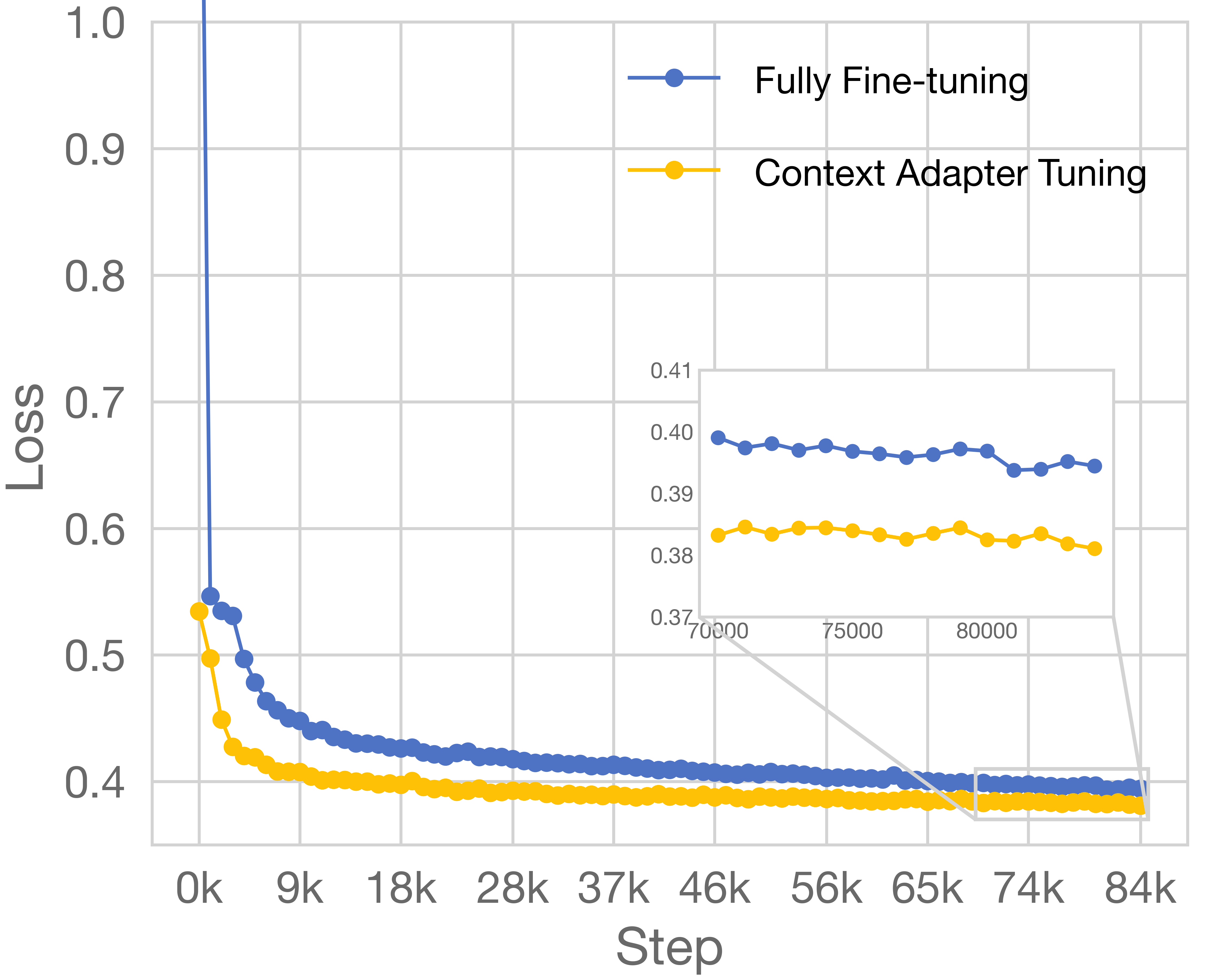}
        \caption{Base structure setting.}
        \label{fig:ablation_1}
    \end{subfigure}
    \hspace{-0.1cm}
    \begin{subfigure}[b]{0.48\columnwidth}
        \includegraphics[width=\textwidth]{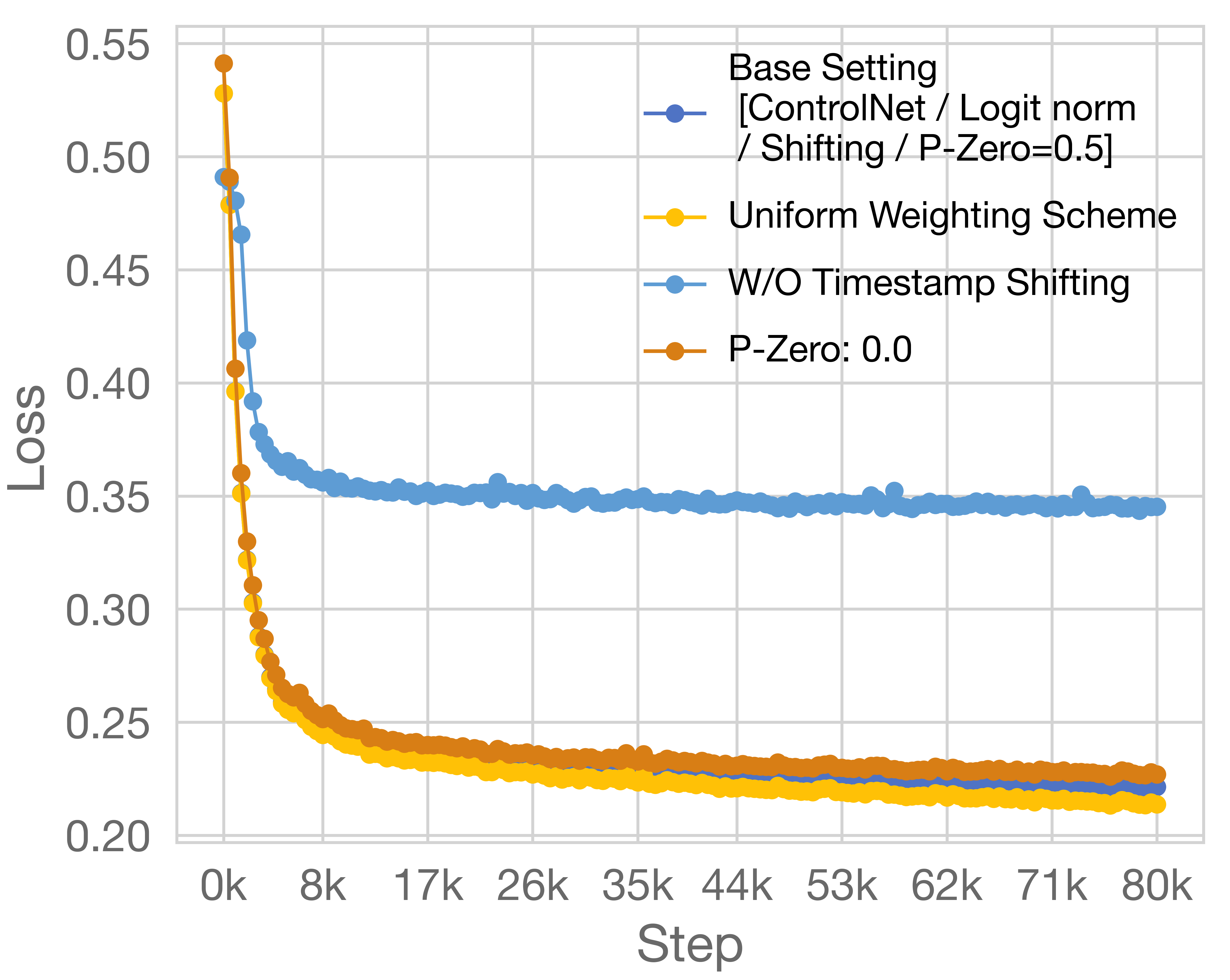}
        \caption{Hyperparameter settings.}
        \label{fig:ablation_2}
    \end{subfigure}
    \vspace{-0.1cm}
    \begin{subfigure}[b]{0.48\columnwidth}
        \includegraphics[width=\textwidth]{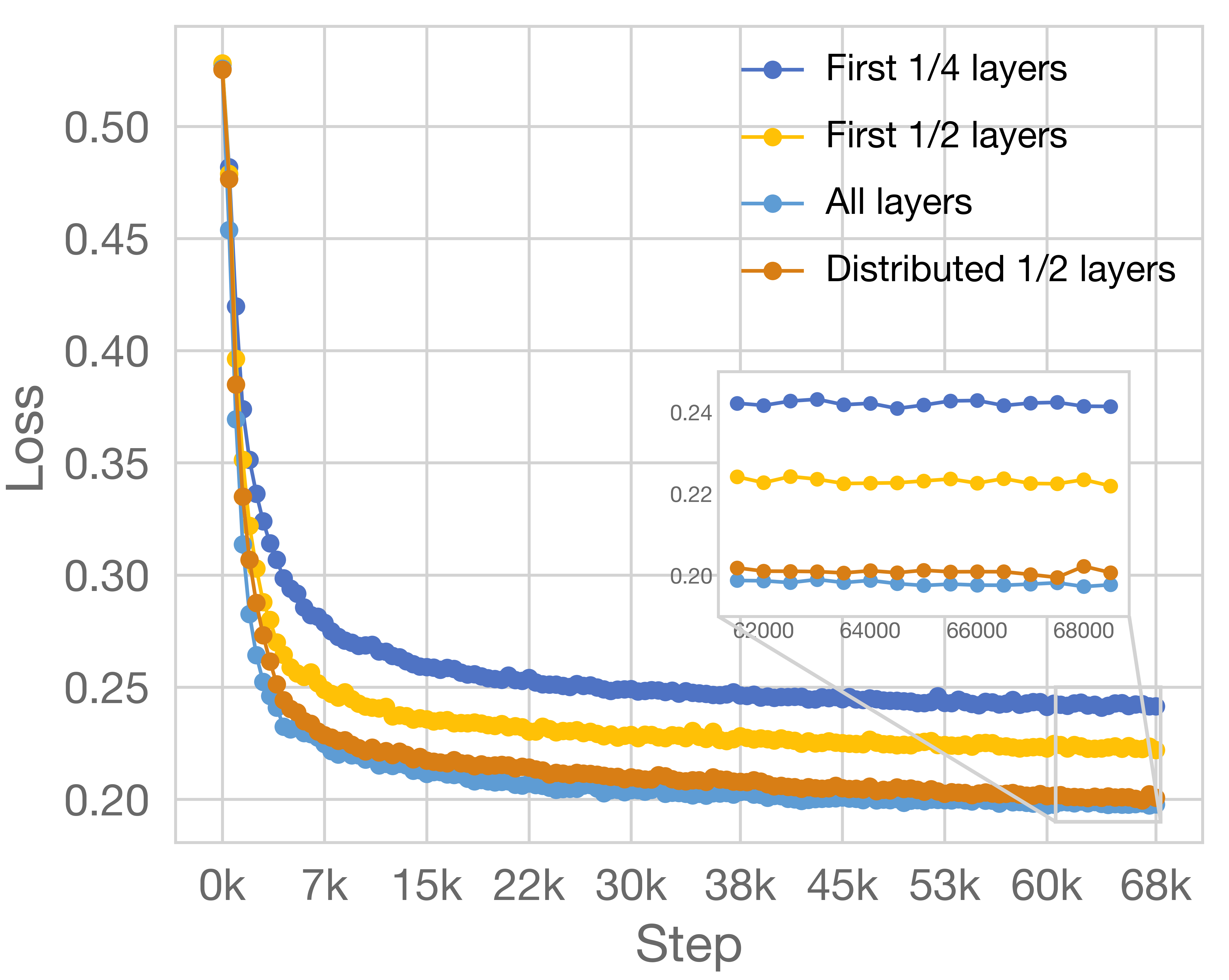}
        \caption{Context adapter configurations.}
        \label{fig:ablation_3}
    \end{subfigure}
    \hspace{-0.1cm}
    \begin{subfigure}[b]{0.48\columnwidth}
        \includegraphics[width=\textwidth]{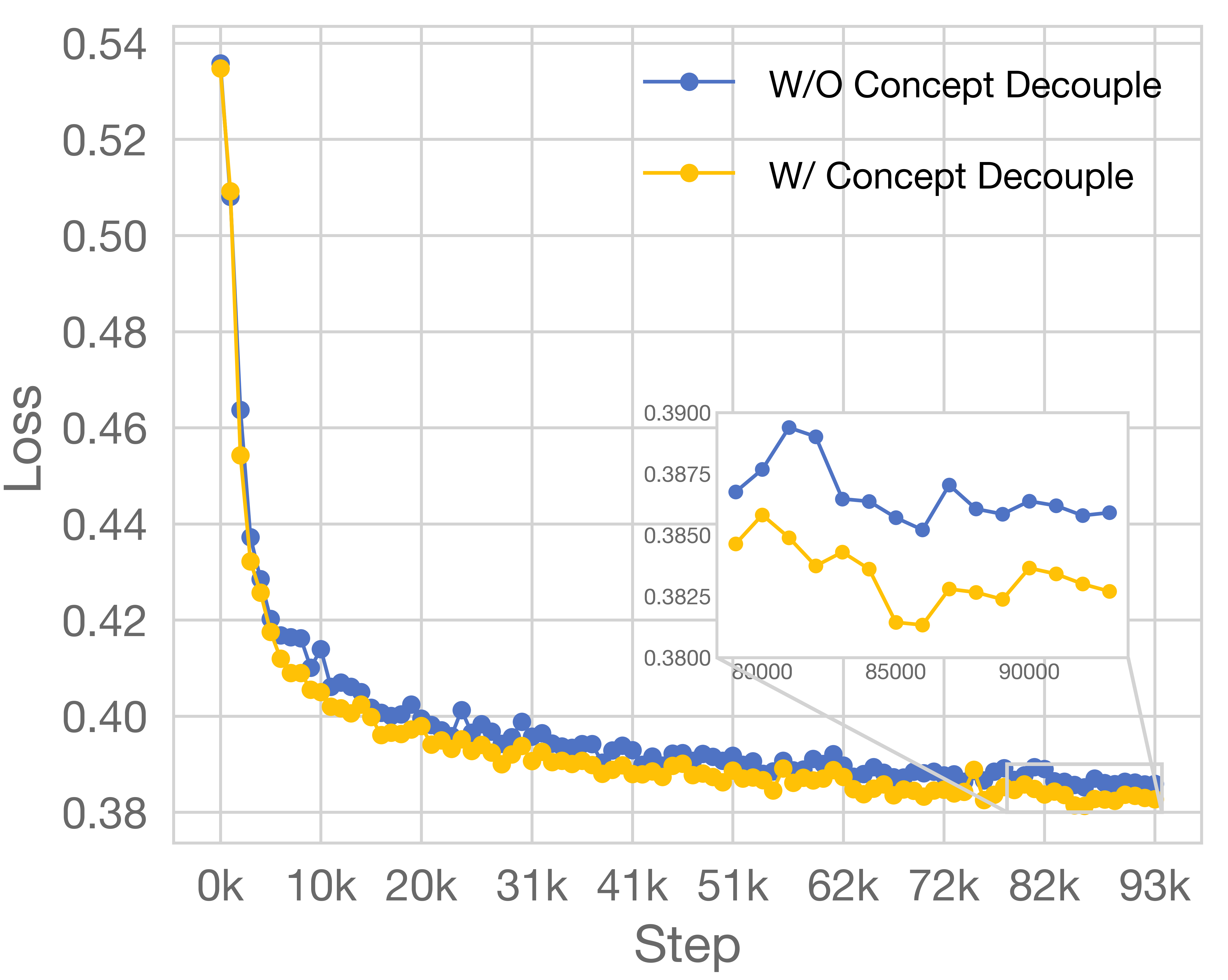}
        \caption{Concept decouple setting.}
        \label{fig:ablation_4}
    \end{subfigure}
    \caption{\textbf{Ablation Studies} of the \method regarding structures, hyperparameters, and module configurations.}
    \label{fig:ablation}
\end{figure}

To better understand the impact of different independent modules on a unified video generation framework, we conducted a series of systematic comparative experiments based on the LTX-Video model to achieve a better model structure and configuration. To accurately assess the different experimental settings, we sample 250 data points for each task as a validation set and calculate the training loss, reflecting the model's training progress through the mean curve changes of different tasks.

\noindent
\textbf{Base Structure.} 
Text-guided image or video generation models only take noise as inference input. When extend to our unified input paradigm, VCU, we can conduct training using fully fine-tuning or by incorporating additional parameter fine-tuning. 
Specifically, as shown in ~\cref{fig:ablation_1}, we compare the concatenation of different inputs along the channel dimension and modify the input dimensions of the patchify projection layer to achieve the loading and fully fine-tuning of the pre-trained model. 
Additionally, we introduce some extra training parameters in the form of Res-Tuning~\cite{restuning}, which serialize VCU in a bypass branch and inject information into the main branch. The results indicate that both methods yielded similar effects; however, since the additional parameter fine-tuning converge faster, we base our subsequent experiments on this approach.
As shown in ~\cref{fig:ablation_2}, we further conduct hyperparameter experiments based on this structure, focusing on aspects such as weighting schemes, timestamp shifting, and p-zero.

\noindent
\textbf{Context Adapter.} 
Since the number of context blocks will significantly effect the model size and inference time consumption, we attempt to find an optimal number and distribution of context blocks.
We begin with selecting continuous blocks at the input side and make comparisons between the first 1/4 blocks, 1/2 blocks, and all blocks. 
Inspired by the Res-Tuning~\cite{restuning} method, we also experiment with evenly distributing the injection blocks instead of selecting a continuous block series. 
As shown in ~\cref{fig:ablation_3}, we can see that when using the same number of blocks, the distributed arrangement of blocks outperforms the continuous arrangement in shallow blocks. 
Furthermore, a greater number of blocks generally yields better results, but due to the limited improvement in effectiveness and the constraints of training resources, we adopt a partially distributed arrangement of blocks.

\noindent
\textbf{Concept Decouple.} 
During training, we introduce a Concept Decouple processing module to further disassemble the visual units, clarifying what content the model needs to learn to modify or retain. As shown in ~\cref{fig:ablation_4}, using this module result in a more significant reduction in loss.

\section{Conclusion}
\label{sec:conclusion}

This paper introduces \method, an all-in-one video generation and editing framework. 
It unifies the diverse and complex multimodal inputs required for various video tasks, bridging the gap between specialized models for each individual task. 
This enables most video AI creation tasks to be completed with a single inference of a single model. 
While broadly covering various video tasks, \method also supports flexible and free combinations of these tasks, greatly expanding the application scenarios of video generation models and meeting a wide range of user creative needs. 
The \method framework paves the way for the development of unified visual generative models with multimodal inputs and represents a significant milestone in the field of visual generation.

\noindent
\textbf{Acknowledgments.} \label{sec:ack}
We would like to express our sincere appreciation for the contributions of many colleagues for their insightful discussions, valuable suggestions, and constructive feedback, including:
Yuwei Wang, Haiming Zhao, Chenwei Xie and Sheng Yao for their data contributions, and Shiwei Zhang, Tao Fang, Xiang Wang for their discussions and suggestions.
{
    \small
    \bibliographystyle{ieeenat_fullname}
    \bibliography{main}

\begin{thebibliography}{82}
\providecommand{\natexlab}[1]{#1}
\providecommand{\url}[1]{\texttt{#1}}
\expandafter\ifx\csname urlstyle\endcsname\relax
  \providecommand{\doi}[1]{doi: #1}\else
  \providecommand{\doi}{doi: \begingroup \urlstyle{rm}\Url}\fi

\bibitem[AI(2025)]{klingai}
KLING AI.
\newblock {KLING AI, \url{https://klingai.com/}}, 2025.

\bibitem[AI(2022{\natexlab{a}})]{sd15}
Runway AI.
\newblock {Stable Diffusion v1.5 Model Card, \url{https://huggingface.co/runwayml/stable-diffusion-v1-5}}, 2022{\natexlab{a}}.

\bibitem[AI(2022{\natexlab{b}})]{sdinp}
Runway AI.
\newblock {Stable Diffusion Inpainting Model Card, \url{https://huggingface.co/runwayml/stable-diffusion-inpainting}}, 2022{\natexlab{b}}.

\bibitem[Brooks et~al.(2023)Brooks, Holynski, and Efros]{ip2p}
Tim Brooks, Aleksander Holynski, and Alexei~A. Efros.
\newblock {InstructPix2Pix}: {Learning} {To} {Follow} {Image} {Editing} {Instructions}.
\newblock In \emph{IEEE Conf. Comput. Vis. Pattern Recog.}, pages 18392--18402, 2023.

\bibitem[Cao et~al.(2021)Cao, Hidalgo, Simon, Wei, and Sheikh]{openpose}
Zhe Cao, Gines Hidalgo, Tomas Simon, Shih-En Wei, and Yaser Sheikh.
\newblock {OpenPose}: {Realtime} {Multi}-{Person} {2D} {Pose} {Estimation} {Using} {Part} {Affinity} {Fields}.
\newblock \emph{IEEE Trans. Pattern Anal. Mach. Intell.}, 43\penalty0 (1):\penalty0 172--186, 2021.

\bibitem[Chan et~al.(2022)Chan, Durand, and Isola]{infordraws}
Caroline Chan, Fr\'edo Durand, and Phillip Isola.
\newblock {Learning} {To} {Generate} {Line} {Drawings} {That} {Convey} {Geometry} and {Semantics}.
\newblock In \emph{IEEE Conf. Comput. Vis. Pattern Recog.}, pages 7915--7925, 2022.

\bibitem[Chen et~al.(2023{\natexlab{a}})Chen, Yu, Ge, Yao, Xie, Wu, Wang, Kwok, Luo, Lu, and Li]{pixart}
Junsong Chen, Jincheng Yu, Chongjian Ge, Lewei Yao, Enze Xie, Yue Wu, Zhongdao Wang, James Kwok, Ping Luo, Huchuan Lu, and Zhenguo Li.
\newblock {{PixArt-$\alpha$}}: {{Fast Training}} of {{Diffusion Transformer}} for {{Photorealistic Text-to-Image Synthesis}}.
\newblock \emph{arXiv preprint arXiv:2310.00426}, 2023{\natexlab{a}}.

\bibitem[Chen et~al.(2025{\natexlab{a}})Chen, Ma, Wang, Yuan, Zhao, Tian, Wang, Min, Chen, and Liu]{followyourcanvas}
Qihua Chen, Yue Ma, Hongfa Wang, Junkun Yuan, Wenzhe Zhao, Qi Tian, Hongmei Wang, Shaobo Min, Qifeng Chen, and Wei Liu.
\newblock Follow-{{Your-Canvas}}: {{Higher-Resolution Video Outpainting}} with {{Extensive Content Generation}}.
\newblock In \emph{Assoc. Adv. Artif. Intell.}, 2025{\natexlab{a}}.

\bibitem[Chen et~al.(2025{\natexlab{b}})Chen, Ge, Zhang, Zhang, Zhu, Yang, Hao, Wu, Lai, Hu, Lin, Zhang, Li, Li, Wang, Peng, Sun, Luo, Jiang, Yuan, Peng, and Liu]{goku}
Shoufa Chen, Chongjian Ge, Yuqi Zhang, Yida Zhang, Fengda Zhu, Hao Yang, Hongxiang Hao, Hui Wu, Zhichao Lai, Yifei Hu, Ting-Che Lin, Shilong Zhang, Fu Li, Chuan Li, Xing Wang, Yanghua Peng, Peize Sun, Ping Luo, Yi Jiang, Zehuan Yuan, Bingyue Peng, and Xiaobing Liu.
\newblock Goku: {{Flow Based Video Generative Foundation Models}}.
\newblock \emph{arXiv preprint arXiv:2502.04896}, 2025{\natexlab{b}}.

\bibitem[Chen et~al.(2023{\natexlab{b}})Chen, Ji, Wu, Wu, Xie, Li, Xia, Xiao, and Lin]{controlavideo}
Weifeng Chen, Yatai Ji, Jie Wu, Hefeng Wu, Pan Xie, Jiashi Li, Xin Xia, Xuefeng Xiao, and Liang Lin.
\newblock Control-{{A-Video}}: {{Controllable Text-to-Video Diffusion Models}} with {{Motion Prior}} and {{Reward Feedback Learning}}.
\newblock \emph{arXiv preprint arXiv:2305.13840}, 2023{\natexlab{b}}.

\bibitem[Chen et~al.(2023{\natexlab{c}})Chen, Huang, Liu, Shen, Zhao, and Zhao]{anydoor}
Xi Chen, Lianghua Huang, Yu Liu, Yujun Shen, Deli Zhao, and Hengshuang Zhao.
\newblock {{AnyDoor}}: {{Zero-shot Object-level Image Customization}}.
\newblock \emph{arXiv preprint arXiv:2307.09481}, 2023{\natexlab{c}}.

\bibitem[Chen et~al.(2024)Chen, Zhang, Zhang, Zhou, Kim, Liu, Li, Zhang, Zhao, Wang, Ding, Lin, and Zhao]{unireal}
Xi Chen, Zhifei Zhang, He Zhang, Yuqian Zhou, Soo~Ye Kim, Qing Liu, Yijun Li, Jianming Zhang, Nanxuan Zhao, Yilin Wang, Hui Ding, Zhe Lin, and Hengshuang Zhao.
\newblock {{UniReal}}: {{Universal Image Generation}} and {{Editing}} via {{Learning Real-world Dynamics}}.
\newblock \emph{arXiv preprint arXiv:2412.07774}, 2024.

\bibitem[Cloud(2023)]{wanx}
Alibaba Cloud.
\newblock {Tongyi Wanxiang, \url{https://tongyi.aliyun.com/wanxiang}}, 2023.

\bibitem[Cong et~al.(2024)Cong, Xu, Simon, Chen, Ren, Xie, {Perez-Rua}, Rosenhahn, Xiang, and He]{flatten}
Yuren Cong, Mengmeng Xu, Christian Simon, Shoufa Chen, Jiawei Ren, Yanping Xie, Juan-Manuel {Perez-Rua}, Bodo Rosenhahn, Tao Xiang, and Sen He.
\newblock {{FLATTEN}}: Optical {{FLow-guided ATTENtion}} for consistent text-to-video editing.
\newblock In \emph{Int. Conf. Learn. Represent.}, 2024.

\bibitem[Duan et~al.(2024)Duan, Zhao, Yan, Li, Chen, Xu, Luo, Zhang, Gong, and Xia]{unicadapter}
Lunhao Duan, Shanshan Zhao, Wenjun Yan, Yinglun Li, Qing-Guo Chen, Zhao Xu, Weihua Luo, Kaifu Zhang, Mingming Gong, and Gui-Song Xia.
\newblock {{UNIC-Adapter}}: {{Unified Image-instruction Adapter}} with {{Multi-modal Transformer}} for {{Image Generation}}.
\newblock \emph{arXiv preprint arXiv:2412.18928}, 2024.

\bibitem[Esser et~al.(2024)Esser, Kulal, Blattmann, Entezari, Müller, Saini, Levi, Lorenz, Sauer, Boesel, Podell, Dockhorn, English, Lacey, Goodwin, Marek, and Rombach]{sd3}
Patrick Esser, Sumith Kulal, Andreas Blattmann, Rahim Entezari, Jonas Müller, Harry Saini, Yam Levi, Dominik Lorenz, Axel Sauer, Frederic Boesel, Dustin Podell, Tim Dockhorn, Zion English, Kyle Lacey, Alex Goodwin, Yannik Marek, and Robin Rombach.
\newblock Scaling {Rectified} {Flow} {Transformers} for {High}-{Resolution} {Image} {Synthesis}.
\newblock In \emph{Int. Conf. Mach. Learn.}, 2024.

\bibitem[Fan et~al.(2023)Fan, Guo, Gong, Wang, Ge, Jiang, Luo, and Zhan]{m3ddm}
Fanda Fan, Chaoxu Guo, Litong Gong, Biao Wang, Tiezheng Ge, Yuning Jiang, Chunjie Luo, and Jianfeng Zhan.
\newblock Hierarchical {{Masked 3D Diffusion Model}} for {{Video Outpainting}}.
\newblock In \emph{ACM Int. Conf. Multimedia}, pages 7890--7900, 2023.

\bibitem[FLUX(2024)]{flux}
FLUX.
\newblock {FLUX, \url{https://blackforestlabs.ai/}}, 2024.

\bibitem[Ge et~al.(2024)Ge, Zhao, Li, Ge, and Shan]{seededit}
Yuying Ge, Sijie Zhao, Chen Li, Yixiao Ge, and Ying Shan.
\newblock {SEED}-{Data}-{Edit} {Technical} {Report}: {A} {Hybrid} {Dataset} for {Instructional} {Image} {Editing}.
\newblock \emph{arXiv preprint arXiv:2405.04007}, 2024.

\bibitem[Guo et~al.(2024{\natexlab{a}})Guo, Zheng, Hou, Gao, Deng, Wan, Zhang, Liu, Hu, Zha, Huang, and Ma]{i2vadapter}
Xun Guo, Mingwu Zheng, Liang Hou, Yuan Gao, Yufan Deng, Pengfei Wan, Di Zhang, Yufan Liu, Weiming Hu, Zhengjun Zha, Haibin Huang, and Chongyang Ma.
\newblock {{I2V-Adapter}}: {{A General Image-to-Video Adapter}} for {{Diffusion Models}}.
\newblock In \emph{ACM SIGGRAPH}, pages 1--12, 2024{\natexlab{a}}.

\bibitem[Guo et~al.(2024{\natexlab{b}})Guo, Wu, Chen, Chen, Zhang, and He]{pulid}
Zinan Guo, Yanze Wu, Zhuowei Chen, Lang Chen, Peng Zhang, and Qian He.
\newblock {PuLID}: {Pure} and {Lightning} {ID} {Customization} via {Contrastive} {Alignment}.
\newblock In \emph{Adv. Neural Inform. Process. Syst.}, 2024{\natexlab{b}}.

\bibitem[HaCohen et~al.(2025)HaCohen, Chiprut, Brazowski, Shalem, Moshe, Richardson, Levin, Shiran, Zabari, Gordon, Panet, Weissbuch, Kulikov, Bitterman, Melumian, and Bibi]{ltx}
Yoav HaCohen, Nisan Chiprut, Benny Brazowski, Daniel Shalem, Dudu Moshe, Eitan Richardson, Eran Levin, Guy Shiran, Nir Zabari, Ori Gordon, Poriya Panet, Sapir Weissbuch, Victor Kulikov, Yaki Bitterman, Zeev Melumian, and Ofir Bibi.
\newblock {{LTX-Video}}: {{Realtime Video Latent Diffusion}}.
\newblock \emph{arXiv preprint arXiv:2501.00103}, 2025.

\bibitem[Han et~al.(2025)Han, Jiang, Pan, Zhang, Mao, Xie, Liu, and Zhou]{ace}
Zhen Han, Zeyinzi Jiang, Yulin Pan, Jingfeng Zhang, Chaojie Mao, Chenwei Xie, Yu Liu, and Jingren Zhou.
\newblock {{ACE}}: {{All-round Creator}} and {{Editor Following Instructions}} via {{Diffusion Transformer}}.
\newblock In \emph{Int. Conf. Learn. Represent.}, 2025.

\bibitem[Ho and Salimans(2021)]{cfg}
Jonathan Ho and Tim Salimans.
\newblock Classifier-{Free} {Diffusion} {Guidance}.
\newblock In \emph{Adv. Neural Inform. Process. Syst.}, 2021.

\bibitem[Ho et~al.(2020)Ho, Jain, and Abbeel]{ddpm}
Jonathan Ho, Ajay Jain, and Pieter Abbeel.
\newblock Denoising {Diffusion} {Probabilistic} {Models}.
\newblock In \emph{Adv. Neural Inform. Process. Syst.} Curran Associates, Inc., 2020.

\bibitem[Huang et~al.(2023)Huang, Chen, Liu, Shen, Zhao, and Zhou]{composer}
Lianghua Huang, Di Chen, Yu Liu, Yujun Shen, Deli Zhao, and Jingren Zhou.
\newblock Composer: {Creative} and {Controllable} {Image} {Synthesis} with {Composable} {Conditions}.
\newblock In \emph{Int. Conf. Mach. Learn.}, 2023.

\bibitem[Huang et~al.(2024{\natexlab{a}})Huang, He, Yu, Zhang, Si, Jiang, Zhang, Wu, Jin, Chanpaisit, Wang, Chen, Wang, Lin, Qiao, and Liu]{vbench}
Ziqi Huang, Yinan He, Jiashuo Yu, Fan Zhang, Chenyang Si, Yuming Jiang, Yuanhan Zhang, Tianxing Wu, Qingyang Jin, Nattapol Chanpaisit, Yaohui Wang, Xinyuan Chen, Limin Wang, Dahua Lin, Yu Qiao, and Ziwei Liu.
\newblock {{VBench}}: {{Comprehensive Benchmark Suite}} for {{Video Generative Models}}.
\newblock In \emph{IEEE Conf. Comput. Vis. Pattern Recog.}, pages 21807--21818, 2024{\natexlab{a}}.

\bibitem[Huang et~al.(2024{\natexlab{b}})Huang, Zhang, Xu, He, Yu, Dong, Ma, Chanpaisit, Si, Jiang, Wang, Chen, Chen, Wang, Lin, Qiao, and Liu]{vbenchpp}
Ziqi Huang, Fan Zhang, Xiaojie Xu, Yinan He, Jiashuo Yu, Ziyue Dong, Qianli Ma, Nattapol Chanpaisit, Chenyang Si, Yuming Jiang, Yaohui Wang, Xinyuan Chen, Ying-Cong Chen, Limin Wang, Dahua Lin, Yu Qiao, and Ziwei Liu.
\newblock {{VBench}}++: {{Comprehensive}} and {{Versatile Benchmark Suite}} for {{Video Generative Models}}.
\newblock \emph{arXiv preprint arXiv:2411.13503}, 2024{\natexlab{b}}.

\bibitem[Jiang et~al.(2023)Jiang, Mao, Huang, Ma, Lv, Shen, Zhao, and Zhou]{restuning}
Zeyinzi Jiang, Chaojie Mao, Ziyuan Huang, Ao Ma, Yiliang Lv, Yujun Shen, Deli Zhao, and Jingren Zhou.
\newblock Res-{Tuning}: {A} {Flexible} and {Efficient} {Tuning} {Paradigm} via {Unbinding} {Tuner} from {Backbone}.
\newblock In \emph{Adv. Neural Inform. Process. Syst.}, 2023.

\bibitem[Jiang et~al.(2024)Jiang, Mao, Pan, Han, and Zhang]{scedit}
Zeyinzi Jiang, Chaojie Mao, Yulin Pan, Zhen Han, and Jingfeng Zhang.
\newblock {{SCEdit}}: {{Efficient}} and {{Controllable Image Diffusion Generation}} via {{Skip Connection Editing}}.
\newblock In \emph{IEEE Conf. Comput. Vis. Pattern Recog.}, pages 8995--9004, 2024.

\bibitem[Khachatryan et~al.(2023)Khachatryan, Movsisyan, Tadevosyan, Henschel, Wang, Navasardyan, and Shi]{text2videozero}
Levon Khachatryan, Andranik Movsisyan, Vahram Tadevosyan, Roberto Henschel, Zhangyang Wang, Shant Navasardyan, and Humphrey Shi.
\newblock {{Text2Video-Zero}}: {{Text-to-Image Diffusion Models}} are {{Zero-Shot Video Generators}}.
\newblock In \emph{Int. Conf. Comput. Vis.}, pages 15954--15964, 2023.

\bibitem[Kong et~al.(2024)Kong, Tian, Zhang, Min, Dai, Zhou, Xiong, Li, Wu, Zhang, Wu, Lin, Yuan, Long, Wang, Wang, Li, Huang, Yang, Tan, Wang, Song, Bai, Wu, Xue, Wang, Wang, Liu, Li, Li, Wang, Yu, Deng, Li, Chen, Cui, Peng, Yu, He, Xu, Zhou, Xu, Tao, Lu, Liu, Zhou, Wang, Yang, Wang, Liu, Jiang, and Zhong]{hunyuanvideo}
Weijie Kong, Qi Tian, Zijian Zhang, Rox Min, Zuozhuo Dai, Jin Zhou, Jiangfeng Xiong, Xin Li, Bo Wu, Jianwei Zhang, Kathrina Wu, Qin Lin, Junkun Yuan, Yanxin Long, Aladdin Wang, Andong Wang, Changlin Li, Duojun Huang, Fang Yang, Hao Tan, Hongmei Wang, Jacob Song, Jiawang Bai, Jianbing Wu, Jinbao Xue, Joey Wang, Kai Wang, Mengyang Liu, Pengyu Li, Shuai Li, Weiyan Wang, Wenqing Yu, Xinchi Deng, Yang Li, Yi Chen, Yutao Cui, Yuanbo Peng, Zhentao Yu, Zhiyu He, Zhiyong Xu, Zixiang Zhou, Zunnan Xu, Yangyu Tao, Qinglin Lu, Songtao Liu, Dax Zhou, Hongfa Wang, Yong Yang, Di Wang, Yuhong Liu, Jie Jiang, and Caesar Zhong.
\newblock {{HunyuanVideo}}: {{A Systematic Framework For Large Video Generative Models}}.
\newblock \emph{arXiv preprint arXiv:2412.03603}, 2024.

\bibitem[Li et~al.(2024)Li, Zhang, Lin, Xiong, Long, Deng, Zhang, Liu, Huang, Xiao, et~al.]{hunyuan_dit}
Zhimin Li, Jianwei Zhang, Qin Lin, Jiangfeng Xiong, Yanxin Long, Xinchi Deng, Yingfang Zhang, Xingchao Liu, Minbin Huang, Zedong Xiao, et~al.
\newblock Hunyuan-{DiT}: {A} {Powerful} {Multi}-{Resolution} {Diffusion} {Transformer} with {Fine}-{Grained} {Chinese} {Understanding}.
\newblock \emph{arXiv preprint arXiv:2405.08748}, 2024.

\bibitem[Liew et~al.(2023)Liew, Yan, Zhang, Xu, and Feng]{magicedit}
Jun~Hao Liew, Hanshu Yan, Jianfeng Zhang, Zhongcong Xu, and Jiashi Feng.
\newblock {{MagicEdit}}: {{High-Fidelity}} and {{Temporally Coherent Video Editing}}.
\newblock \emph{arXiv preprint arXiv:2308.14749}, 2023.

\bibitem[Liu et~al.(2025)Liu, Ma, Li, Chen, Liu, He, and Wu]{phantom}
Lijie Liu, Tianxiang Ma, Bingchuan Li, Zhuowei Chen, Jiawei Liu, Qian He, and Xinglong Wu.
\newblock Phantom: {{Subject-consistent}} video generation via cross-modal alignment.
\newblock \emph{arXiv preprint arXiv:2502.11079}, 2025.

\bibitem[Liu et~al.(2023{\natexlab{a}})Liu, Zeng, Ren, Li, Zhang, Yang, Li, Yang, Su, Zhu, and Zhang]{groundingdino}
Shilong Liu, Zhaoyang Zeng, Tianhe Ren, Feng Li, Hao Zhang, Jie Yang, Chunyuan Li, Jianwei Yang, Hang Su, Jun Zhu, and Lei Zhang.
\newblock Grounding {{DINO}}: {{Marrying DINO}} with {{Grounded Pre-Training}} for {{Open-Set Object Detection}}.
\newblock \emph{arXiv preprint arXiv:2303.05499}, 2023{\natexlab{a}}.

\bibitem[Liu et~al.(2024)Liu, Zhang, Li, Lin, and Jia]{videop2p}
Shaoteng Liu, Yuechen Zhang, Wenbo Li, Zhe Lin, and Jiaya Jia.
\newblock Video-{{P2P}}: {{Video Editing}} with {{Cross-attention Control}}.
\newblock In \emph{IEEE Conf. Comput. Vis. Pattern Recog.}, pages 8599--8608, 2024.

\bibitem[Liu et~al.(2023{\natexlab{b}})Liu, Feng, Zhu, Zhang, Zheng, Liu, Zhao, Zhou, and Cao]{cones}
Zhiheng Liu, Ruili Feng, Kai Zhu, Yifei Zhang, Kecheng Zheng, Yu Liu, Deli Zhao, Jingren Zhou, and Yang Cao.
\newblock Cones: {{Concept Neurons}} in {{Diffusion Models}} for {{Customized Generation}}.
\newblock In \emph{Int. Conf. Mach. Learn.}, 2023{\natexlab{b}}.

\bibitem[Liu et~al.(2023{\natexlab{c}})Liu, Zhang, Shen, Zheng, Zhu, Feng, Liu, Zhao, Zhou, and Cao]{cones2}
Zhiheng Liu, Yifei Zhang, Yujun Shen, Kecheng Zheng, Kai Zhu, Ruili Feng, Yu Liu, Deli Zhao, Jingren Zhou, and Yang Cao.
\newblock Cones 2: {{Customizable Image Synthesis}} with {{Multiple Subjects}}.
\newblock In \emph{Adv. Neural Inform. Process. Syst.}, 2023{\natexlab{c}}.

\bibitem[Ma et~al.(2024)Ma, He, Cun, Wang, Chen, Shan, Li, and Chen]{followyourpose}
Yue Ma, Yingqing He, Xiaodong Cun, Xintao Wang, Siran Chen, Ying Shan, Xiu Li, and Qifeng Chen.
\newblock Follow {{Your Pose}}: {{Pose-Guided Text-to-Video Generation}} using {{Pose-Free Videos}}.
\newblock In \emph{Assoc. Adv. Artif. Intell.}, 2024.

\bibitem[Mao et~al.(2025)Mao, Zhang, Pan, Jiang, Han, Liu, and Zhou]{acepp}
Chaojie Mao, Jingfeng Zhang, Yulin Pan, Zeyinzi Jiang, Zhen Han, Yu Liu, and Jingren Zhou.
\newblock {{ACE}}++: {{Instruction-Based Image Creation}} and {{Editing}} via {{Context-Aware Content Filling}}.
\newblock \emph{arXiv preprint arXiv:2501.02487}, 2025.

\bibitem[Meng et~al.(2021)Meng, He, Song, Song, Wu, Zhu, and Ermon]{sdedit}
Chenlin Meng, Yutong He, Yang Song, Jiaming Song, Jiajun Wu, Jun-Yan Zhu, and Stefano Ermon.
\newblock {SDEdit}: {Guided} {Image} {Synthesis} and {Editing} with {Stochastic} {Differential} {Equations}.
\newblock In \emph{Int. Conf. Learn. Represent.}, 2021.

\bibitem[Midjourney(2023)]{midjourney}
Midjourney.
\newblock {Midjourney, \url{https://www.midjourney.com}}, 2023.

\bibitem[MiniMax(2024)]{hailuoai}
MiniMax.
\newblock {Hailuo AI Video, \url{https://hailuoai.com/video}}, 2024.

\bibitem[OpenAI(2023)]{dalle3}
OpenAI.
\newblock {DALL·E 3, \url{https://openai.com/dall-e-3}}, 2023.

\bibitem[Pan et~al.(2023)Pan, Tewari, Leimk{\"u}hler, Liu, Meka, and Theobalt]{draggan}
Xingang Pan, Ayush Tewari, Thomas Leimk{\"u}hler, Lingjie Liu, Abhimitra Meka, and Christian Theobalt.
\newblock Drag {{Your GAN}}: {{Interactive Point-based Manipulation}} on the {{Generative Image Manifold}}.
\newblock In \emph{ACM SIGGRAPH}, 2023.

\bibitem[Pan et~al.(2024)Pan, Mao, Jiang, Han, and Zhang]{largen}
Yulin Pan, Chaojie Mao, Zeyinzi Jiang, Zhen Han, and Jingfeng Zhang.
\newblock Locate, {Assign}, {Refine}: {Taming} {Customized} {Image} {Inpainting} with {Text}-{Subject} {Guidance}.
\newblock \emph{arXiv preprint arXiv:2403.19534}, 2024.

\bibitem[Peebles and Xie(2023)]{dit}
William Peebles and Saining Xie.
\newblock Scalable {Diffusion} {Models} with {Transformers}.
\newblock In \emph{Int. Conf. Comput. Vis.}, pages 4195--4305, 2023.

\bibitem[PiKa(2025)]{pika}
PiKa.
\newblock {PiKa, \url{https://pika.art/}}, 2025.

\bibitem[Qin et~al.(2023)Qin, Zhang, Yu, Feng, Yang, Zhou, Wang, Niebles, Xiong, Savarese, Ermon, Fu, and Xu]{unicontrol}
Can Qin, Shu Zhang, Ning Yu, Yihao Feng, Xinyi Yang, Yingbo Zhou, Huan Wang, Juan~Carlos Niebles, Caiming Xiong, Silvio Savarese, Stefano Ermon, Yun Fu, and Ran Xu.
\newblock {UniControl}: {A} {Unified} {Diffusion} {Model} for {Controllable} {Visual} {Generation} {In} the {Wild}.
\newblock In \emph{Adv. Neural Inform. Process. Syst.}, 2023.

\bibitem[Ranftl et~al.(2022)Ranftl, Lasinger, Hafner, Schindler, and Koltun]{midas}
René Ranftl, Katrin Lasinger, David Hafner, Konrad Schindler, and Vladlen Koltun.
\newblock Towards {Robust} {Monocular} {Depth} {Estimation}: {Mixing} {Datasets} for {Zero}-{Shot} {Cross}-{Dataset} {Transfer}.
\newblock \emph{IEEE Trans. Pattern Anal. Mach. Intell.}, pages 1623--1637, 2022.

\bibitem[Ravi et~al.(2025)Ravi, Gabeur, Hu, Hu, Ryali, Ma, Khedr, R{\"a}dle, Rolland, Gustafson, Mintun, Pan, Alwala, Carion, Wu, Girshick, Doll{\'a}r, and Feichtenhofer]{sam2}
Nikhila Ravi, Valentin Gabeur, Yuan-Ting Hu, Ronghang Hu, Chaitanya Ryali, Tengyu Ma, Haitham Khedr, Roman R{\"a}dle, Chloe Rolland, Laura Gustafson, Eric Mintun, Junting Pan, Kalyan~Vasudev Alwala, Nicolas Carion, Chao-Yuan Wu, Ross Girshick, Piotr Doll{\'a}r, and Christoph Feichtenhofer.
\newblock {{SAM}} 2: {{Segment Anything}} in {{Images}} and {{Videos}}.
\newblock In \emph{Int. Conf. Learn. Represent.}, 2025.

\bibitem[Rombach et~al.(2022)Rombach, Blattmann, Lorenz, Esser, and Ommer]{ldm}
Robin Rombach, Andreas Blattmann, Dominik Lorenz, Patrick Esser, and Bj{\"o}rn Ommer.
\newblock High-resolution image synthesis with latent diffusion models.
\newblock In \emph{IEEE Conf. Comput. Vis. Pattern Recog.}, pages 10684--10695, 2022.

\bibitem[Ronneberger et~al.(2015)Ronneberger, Fischer, and Brox]{unet}
Olaf Ronneberger, Philipp Fischer, and Thomas Brox.
\newblock U-{Net}: {Convolutional} {Networks} for {Biomedical} {Image} {Segmentation}.
\newblock \emph{Med. Image Comput. Computer-Assisted Interv.}, 2015.

\bibitem[Runway(2025)]{gen3}
Runway.
\newblock {Gen-3, \url{https://app.runwayml.com/video-tools}}, 2025.

\bibitem[Song et~al.(2021{\natexlab{a}})Song, Meng, and Ermon]{ddim}
Jiaming Song, Chenlin Meng, and Stefano Ermon.
\newblock Denoising {Diffusion} {Implicit} {Models}.
\newblock In \emph{Int. Conf. Learn. Represent.}, 2021{\natexlab{a}}.

\bibitem[Song et~al.(2021{\natexlab{b}})Song, Sohl-Dickstein, Kingma, Kumar, Ermon, and Poole]{sde}
Yang Song, Jascha Sohl-Dickstein, Diederik~P. Kingma, Abhishek Kumar, Stefano Ermon, and Ben Poole.
\newblock Score-{Based} {Generative} {Modeling} through {Stochastic} {Differential} {Equations}.
\newblock In \emph{Int. Conf. Learn. Represent.}, 2021{\natexlab{b}}.

\bibitem[StabilityAI(2022{\natexlab{a}})]{sd21}
StabilityAI.
\newblock {Stable Diffusion v2-1 Model Card, \url{https://huggingface.co/stabilityai/stable-diffusion-2-1}}, 2022{\natexlab{a}}.

\bibitem[StabilityAI(2022{\natexlab{b}})]{sdxl}
StabilityAI.
\newblock {Stable Diffusion XL Model Card, \url{https://huggingface.co/stabilityai/stable-diffusion-xl-base-1.0}}, 2022{\natexlab{b}}.

\bibitem[StabilityAI(2024)]{cosxl}
StabilityAI.
\newblock {CosXL Model Card, \url{https://huggingface.co/stabilityai/cosxl}}, 2024.

\bibitem[Sun et~al.(2023)Sun, Yang, Peng, Shen, Yang, Hu, Qiu, and Koike]{imagebrush}
Ya~Sheng Sun, Yifan Yang, Houwen Peng, Yifei Shen, Yuqing Yang, Han Hu, Lili Qiu, and Hideki Koike.
\newblock {ImageBrush}: {Learning} {Visual} {In}-{Context} {Instructions} for {Exemplar}-{Based} {Image} {Manipulation}.
\newblock In \emph{Adv. Neural Inform. Process. Syst.}, 2023.

\bibitem[Suvorov et~al.(2022)Suvorov, Logacheva, Mashikhin, Remizova, Ashukha, Silvestrov, Kong, Goka, Park, and Lempitsky]{lama}
Roman Suvorov, Elizaveta Logacheva, Anton Mashikhin, Anastasia Remizova, Arsenii Ashukha, Aleksei Silvestrov, Naejin Kong, Harshith Goka, Kiwoong Park, and Victor Lempitsky.
\newblock Resolution-{Robust} {Large} {Mask} {Inpainting} {With} {Fourier} {Convolutions}.
\newblock In \emph{IEEE Winter Conf. Appl. Comput. Vis.}, pages 2149--2159, 2022.

\bibitem[Tan et~al.(2024)Tan, Liu, Yang, Xue, and Wang]{ominictr}
Zhenxiong Tan, Songhua Liu, Xingyi Yang, Qiaochu Xue, and Xinchao Wang.
\newblock {{OminiControl}}: {{Minimal}} and {{Universal Control}} for {{Diffusion Transformer}}.
\newblock \emph{arXiv preprint arXiv:2411.15098}, 2024.

\bibitem[Team(2025)]{wan2.1}
Wan Team.
\newblock Wan: Open and advanced large-scale video generative models.
\newblock 2025.

\bibitem[Teed and Deng(2020)]{raft}
Zachary Teed and Jia Deng.
\newblock {{RAFT}}: {{Recurrent All-Pairs Field Transforms}} for {{Optical Flow}}.
\newblock In \emph{Eur. Conf. Comput. Vis.}, pages 402--419, 2020.

\bibitem[Vidu(2025)]{vidu}
Vidu.
\newblock {Vidu, \url{https://www.vidu.cn/}}, 2025.

\bibitem[Wang et~al.(2024{\natexlab{a}})Wang, Bai, Wang, Qin, and Chen]{instantid}
Qixun Wang, Xu Bai, Haofan Wang, Zekui Qin, and Anthony Chen.
\newblock {{InstantID}}: {{Zero-shot Identity-Preserving Generation}} in {{Seconds}}.
\newblock \emph{arXiv preprint arXiv:2401.07519}, 2024{\natexlab{a}}.

\bibitem[Wang et~al.(2023)Wang, Yuan, Zhang, Chen, Wang, Zhang, Shen, Zhao, and Zhou]{videocomposer}
Xiang Wang, Hangjie Yuan, Shiwei Zhang, Dayou Chen, Jiuniu Wang, Yingya Zhang, Yujun Shen, Deli Zhao, and Jingren Zhou.
\newblock {{VideoComposer}}: {{Compositional Video Synthesis}} with {{Motion Controllability}}.
\newblock In \emph{Adv. Neural Inform. Process. Syst.}, 2023.

\bibitem[Wang et~al.(2024{\natexlab{b}})Wang, Yuan, Wang, Chen, Xia, Luo, and Shan]{motionctrl}
Zhouxia Wang, Ziyang Yuan, Xintao Wang, Tianshui Chen, Menghan Xia, Ping Luo, and Ying Shan.
\newblock {{MotionCtrl}}: {{A Unified}} and {{Flexible Motion Controller}} for {{Video Generation}}.
\newblock In \emph{ACM SIGGRAPH}, pages 1--11, 2024{\natexlab{b}}.

\bibitem[Wei et~al.(2024)Wei, Zhang, Yuan, Wang, Qiu, Zhao, Feng, Liu, Huang, Ye, Zhang, and Shan]{dreamVideo2}
Yujie Wei, Shiwei Zhang, Hangjie Yuan, Xiang Wang, Haonan Qiu, Rui Zhao, Yutong Feng, Feng Liu, Zhizhong Huang, Jiaxin Ye, Yingya Zhang, and Hongming Shan.
\newblock {{DreamVideo-2}}: {{Zero-Shot Subject-Driven Video Customization}} with {{Precise Motion Control}}.
\newblock \emph{arXiv preprint arXiv:2410.13830}, 2024.

\bibitem[Xiao et~al.(2024)Xiao, Wang, Zhou, Yuan, Xing, Yan, Li, Wang, Huang, and Liu]{omnigen}
Shitao Xiao, Yueze Wang, Junjie Zhou, Huaying Yuan, Xingrun Xing, Ruiran Yan, Chaofan Li, Shuting Wang, Tiejun Huang, and Zheng Liu.
\newblock {{OmniGen}}: {{Unified Image Generation}}.
\newblock \emph{arXiv preprint arXiv:2409.11340}, 2024.

\bibitem[Yang et~al.(2023)Yang, Zeng, Yuan, and Li]{dwpose}
Zhendong Yang, Ailing Zeng, Chun Yuan, and Yu Li.
\newblock Effective {{Whole-body Pose Estimation}} with {{Two-stages Distillation}}.
\newblock In \emph{Int. Conf. Comput. Vis.}, pages 4210--4220, 2023.

\bibitem[Yang et~al.(2025)Yang, Teng, Zheng, Ding, Huang, Xu, Yang, Hong, Zhang, Feng, Yin, Gu, Zhang, Wang, Cheng, Liu, Xu, Dong, and Tang]{cogvideox}
Zhuoyi Yang, Jiayan Teng, Wendi Zheng, Ming Ding, Shiyu Huang, Jiazheng Xu, Yuanming Yang, Wenyi Hong, Xiaohan Zhang, Guanyu Feng, Da Yin, Xiaotao Gu, Yuxuan Zhang, Weihan Wang, Yean Cheng, Ting Liu, Bin Xu, Yuxiao Dong, and Jie Tang.
\newblock {{CogVideoX}}: {{Text-to-Video Diffusion Models}} with {{An Expert Transformer}}.
\newblock In \emph{Int. Conf. Learn. Represent.}, 2025.

\bibitem[Yuan et~al.(2025)Yuan, Huang, He, Ge, Shi, Chen, Luo, and Yuan]{consisid}
Shenghai Yuan, Jinfa Huang, Xianyi He, Yunyuan Ge, Yujun Shi, Liuhan Chen, Jiebo Luo, and Li Yuan.
\newblock Identity-{{Preserving Text-to-Video Generation}} by {{Frequency Decomposition}}.
\newblock In \emph{IEEE Conf. Comput. Vis. Pattern Recog.}, 2025.

\bibitem[Zhang et~al.(2023{\natexlab{a}})Zhang, Mo, Chen, Sun, and Su]{magicbrush}
Kai Zhang, Lingbo Mo, Wenhu Chen, Huan Sun, and Yu Su.
\newblock {{MagicBrush}}: {{A Manually Annotated Dataset}} for {{Instruction-Guided Image Editing}}.
\newblock In \emph{Adv. Neural Inform. Process. Syst.}, 2023{\natexlab{a}}.

\bibitem[Zhang et~al.(2023{\natexlab{b}})Zhang, Rao, and Agrawala]{controlnet}
Lvmin Zhang, Anyi Rao, and Maneesh Agrawala.
\newblock Adding {Conditional} {Control} to {Text}-to-{Image} {Diffusion} {Models}.
\newblock In \emph{Int. Conf. Comput. Vis.}, pages 3836--3847, 2023{\natexlab{b}}.

\bibitem[Zhang et~al.(2023{\natexlab{c}})Zhang, Wang, Zhang, Zhao, Yuan, Qin, Wang, Zhao, and Zhou]{i2vgen}
Shiwei Zhang, Jiayu Wang, Yingya Zhang, Kang Zhao, Hangjie Yuan, Zhiwu Qin, Xiang Wang, Deli Zhao, and Jingren Zhou.
\newblock {{I2VGen-XL}}: {{High-Quality Image-to-Video Synthesis}} via {{Cascaded Diffusion Models}}.
\newblock \emph{arXiv preprint arXiv:2311.04145}, 2023{\natexlab{c}}.

\bibitem[Zhang et~al.(2023{\natexlab{d}})Zhang, Huang, Ma, Li, Luo, Xie, Qin, Luo, Li, Liu, Guo, and Zhang]{ram}
Youcai Zhang, Xinyu Huang, Jinyu Ma, Zhaoyang Li, Zhaochuan Luo, Yanchun Xie, Yuzhuo Qin, Tong Luo, Yaqian Li, Shilong Liu, Yandong Guo, and Lei Zhang.
\newblock Recognize {{Anything}}: {{A Strong Image Tagging Model}}.
\newblock \emph{arXiv preprint arXiv:2306.03514}, 2023{\natexlab{d}}.

\bibitem[Zhang et~al.(2024)Zhang, Wei, Jiang, Zhang, Zuo, and Tian]{controlvideo}
Yabo Zhang, Yuxiang Wei, Dongsheng Jiang, Xiaopeng Zhang, Wangmeng Zuo, and Qi Tian.
\newblock {{ControlVideo}}: {{Training-free Controllable Text-to-Video Generation}}.
\newblock In \emph{Int. Conf. Learn. Represent.}, 2024.

\bibitem[Zhang et~al.(2025)Zhang, Liu, Xia, Peng, Yan, Lo, and Jia]{magicmirror}
Yuechen Zhang, Yaoyang Liu, Bin Xia, Bohao Peng, Zexin Yan, Eric Lo, and Jiaya Jia.
\newblock Magic {{Mirror}}: {{ID-Preserved Video Generation}} in {{Video Diffusion Transformers}}.
\newblock \emph{arXiv preprint arXiv:2411.13503}, 2025.

\bibitem[Zhao et~al.(2024)Zhao, Ma, Chen, Si, Wu, An, Yu, Zhang, Li, and Chang]{ultraedit}
Haozhe Zhao, Xiaojian Ma, Liang Chen, Shuzheng Si, Rujie Wu, Kaikai An, Peiyu Yu, Minjia Zhang, Qing Li, and Baobao Chang.
\newblock {UltraEdit}: {Instruction}-based {Fine}-{Grained} {Image} {Editing} at {Scale}.
\newblock \emph{arXiv preprint arXiv:2407.05282v1}, 2024.

\bibitem[Zhou et~al.(2023)Zhou, Li, Chan, and Loy]{propainter}
Shangchen Zhou, Chongyi Li, Kelvin C.~K. Chan, and Chen~Change Loy.
\newblock {{ProPainter}}: {{Improving Propagation}} and {{Transformer}} for {{Video Inpainting}}.
\newblock In \emph{Int. Conf. Comput. Vis.}, pages 10477--10486, 2023.

\end{thebibliography}
}
\clearpage  
\appendix

\noindent
In the supplementary material, we provide more implementation details (\cref{sup:impl}) including the hyperparameters used in training and inference.
Then, we showcase additional comparisons with existing methods and more qualitative results (\cref{sup:addit}). 
Furthermore, we discuss the social impacts and limitations (\cref{sup:disc}).

\section{Implementation Details}
\label{sup:impl}

\subsection{Hyperparameters}

In ~\cref{suptab:hyper}, we provide an overview of the hyperparameters settings and conduct training based on the foundational text-to-video generation models of LTX-Video~\cite{ltx} and Wan-T2V~\cite{wan2.1}.
The former allows for quick inference with limited resources; in an A100 single-card environment, without a dedicated acceleration strategy, it takes about 24 seconds to sample 40 steps for a video of approximately 5 seconds in duration. This meets the needs of general users for video processing.
In contrast, Wan-T2V is a comprehensive performance video generation model that requires relatively more resources for training and inference, but it is capable of producing high-quality visuals and maintaining smooth temporal consistency.

\begin{table*}[ht]
\caption{Hyperparameter selection for LTX-Video-based and Wan-T2V-based \method.}
\setlength\tabcolsep{10pt}
\centering
\begin{tabular}{l|cc}
\toprule
\multirow{2}{*}{\textbf{Config}} & \multicolumn{2}{c}{\textbf{\#Model}} \\
 & LTX-Video-based & Wan-T2V-based \\
 
\midrule
Task & 12 tasks + composition task & 12 tasks + composition task \\
Batch Size / GPU & 1 & 1/8 \\
Accumulate Step & 8 & 1 \\
Optimizer & AdamW & AdamW \\
Weight Decay & 0.1 & 0.1 \\
Learning Rate & 0.0001 & 0.00005 \\
Learning Rate Schedule & Constant & Constant \\
Training Steps & 200,000 & 200,000 \\
Resolution & \textasciitilde 480p & \textasciitilde 720p \\
Shifting & Ture & True \\
Weighting Scheme & uniform & uniform \\
Sequence Length & 4992 & 75600 \\
Num Layers & 28 & 40 \\
Context Adapter & Res-Tuning & Res-Tuning \\ 
Context Layers & [ 0, 2, 4, 6, 8, 10, 12, 14, 16, 18, 20, 22, 24, 26 ] & [0, 5, 10, 15, 20, 25, 30, 35] \\ 
Concept Decouple & Ture & True \\
Pre-trained Model & LTX-Video-2b-v0.9 & Wan2.1-T2V-14B \\

\midrule
Sampler & Flow Euler & Flow Euler \\
Sample Steps & 40 & 25 \\
Guide Scale & 3.0 & 4.0 \\
Generation speed & \textasciitilde 24s & \textasciitilde 260s (8 gpus) \\

\midrule
Device & A100$\times$16 & A100$\times$128 \\
Training Strategy & AMP / DDP / BFloat16 & FSDP / Tensor Parallel / BFloat16 \\

\bottomrule
\end{tabular}
\label{suptab:hyper}
\end{table*}

\section{Additional Results}
\label{sup:addit}

\subsection{More Visualization}

\begin{figure*}[!ht]  
    \centering  
    \includegraphics[width=0.85\textwidth]{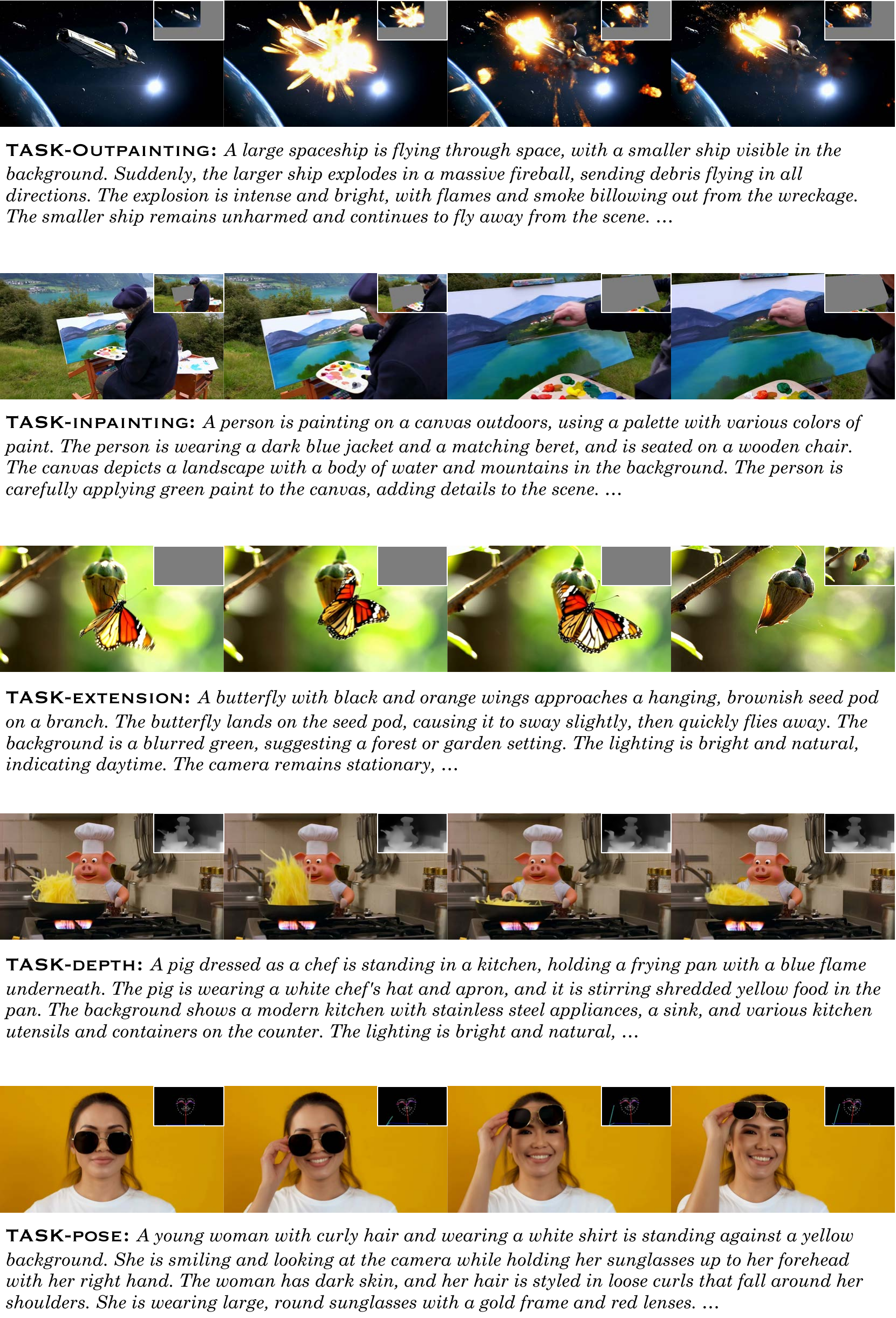}
    \caption{\textbf{More visualization results} of Wan-T2V-based \method framework.} 
    \label{supfig:more_vis_1} %
\end{figure*}

\begin{figure*}[!ht]  
    \centering  
    \includegraphics[width=0.82\textwidth]{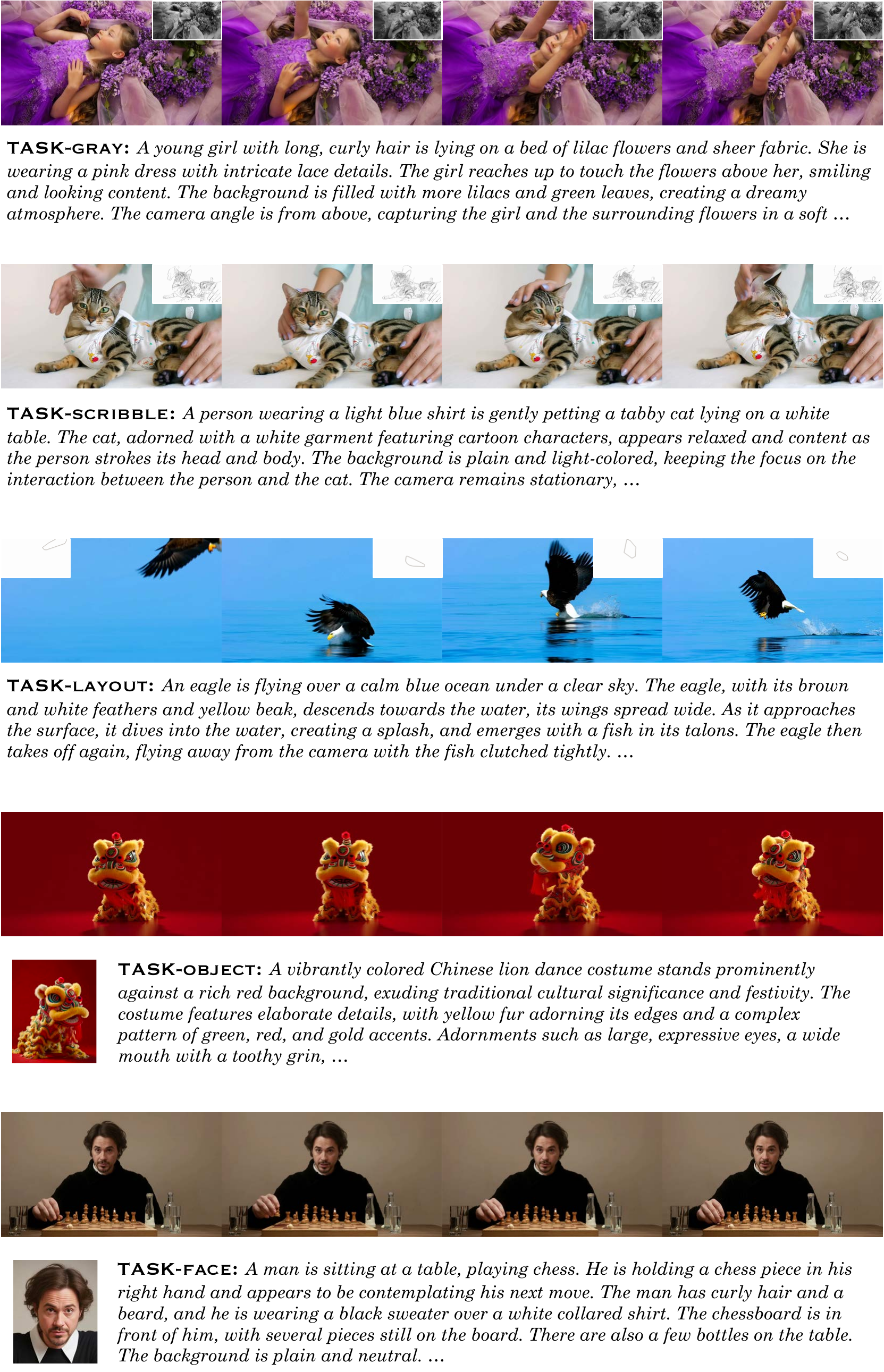}
    \caption{\textbf{More visualization results} of Wan-T2V-based \method framework.} 
    \label{supfig:more_vis_2} %
\end{figure*}

In ~\cref{supfig:more_vis_1} and ~\cref{supfig:more_vis_2}, we present more qualitative results based on Wan-T2V, which include tasks such as outpainting, inpainting, extension, grayscale, depth, scribble, pose, layout, face reference, and object reference.

\subsection{Visualization Comparison}

\begin{figure*}[!ht]  
    \centering  
    \includegraphics[width=0.94\textwidth]{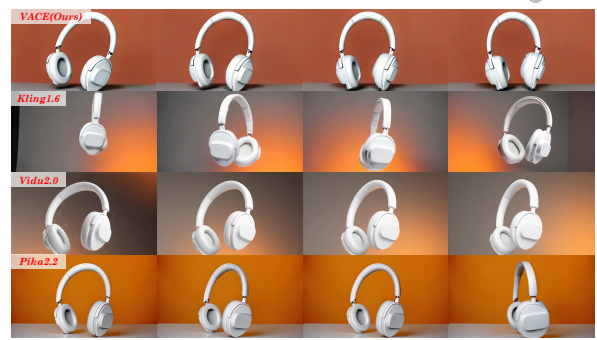}
    \caption{\textbf{Qualitative comparisons} on various tasks based on LTX-Video-based \method framework.} 
    \label{supfig:comp_vis} %
\end{figure*}

In ~\cref{supfig:comp_vis}, we present a visualization of the comparison of the \method based on LTX-Video-2B~\cite{ltx} with others, including the extension task compared with I2VGenXL~\cite{i2vgen}, CogVideoX~\cite{cogvideox}, and LTX-Video-I2V~\cite{ltx}; the unconditional inpainting task compared with ProPainter~\cite{propainter}; the outpainting task with Follow-Your-Canvas~\cite{followyourcanvas} and M3DDM~\cite{m3ddm}; depth-controlled generation with Control-A-Video~\cite{controlavideo}, VideoComposer~\cite{videocomposer}, and ControlVideo~\cite{controlvideo}; pose-controlled generation with Text2Video-Zero~\cite{text2videozero}, ControlVideo~\cite{controlvideo} and Follow-Your-Pose~\cite{followyourpose}; optical flow-controlled generation with FLATTEN~\cite{flatten}; and the reference task compared with commercially closed-source models Keling 1.6~\cite{klingai}, Pika 2.2~\cite{pika}, and Vidu 2.0~\cite{vidu}.

\section{Discussion}
\label{sup:disc}

\subsection{Limitations}
First, the quality of generated content and the overall style are often influenced by the foundation model. This paper verifies this across different model scales: smaller models are advantageous for rapid video generation, but the quality and coherence of the videos are inevitably challenged; larger parameter models significantly improve the success rate of creative output, but the inference speed slows down and resource consumption increases. Finding a relative balance between the two is also a key focus of our future work.

Secondly, compared to the foundational models for text-to-video generation, the current unified models have not been trained on large-scale data and computational power. This results in issues such as the inability to fully maintain identity during reference generation and a lack of complete control over inputs when performing compositional tasks. As discussed in the paper regarding full fine-tuning and additional parameter fine-tuning, when unified tasks begin to apply scaling laws, the results are promising.

In addition, the operational methods for the unified models, compared to image models, present certain challenges due to the inclusion of temporal information and various modalities in their inputs. This aspect creates a threshold for practical usage. Therefore, it is worth exploring how to effectively leverage the capabilities of existing language models or agent models to guide video generation and editing, thereby enhancing productivity.

\subsection{Societal impacts}
From a positive perspective, intelligent video generation and editing can provide creators with a range of innovative tools, helping them to spark new ideas and enhance the artistic and innovative quality of video content.
These technologies are gradually being applied across various industries; for example, in the business sector, video generation technology is transforming marketing and advertising strategies. 
Companies can quickly produce high-quality promotional videos, effectively communicating brand messages and attracting consumers. 
This ability to increase efficiency not only saves labor costs but also enables businesses to implement more creative marketing strategies, thus enhancing their market competitiveness.

However, with the proliferation of these technologies, certain social challenges have emerged. 
The convenience of video generation and editing may lead to the spread of misinformation and false content, undermining the public's trust in information. 
Additionally, when generating content, the technology may inadvertently reinforce existing biases and stereotypes, negatively impacting societal cultural perceptions. 
These issues prompt reflections on ethics and responsibility, calling for policymakers, technology developers, and various sectors of society to work together to establish appropriate regulations to ensure the healthy development of these technologies. 
We must also examine their potential impacts with a cautious attitude, actively exploring ways to balance innovation with social responsibility, so that they can deliver greater benefits to society.

\end{document}